\pdfoutput=1
\documentclass[10pt,twocolumn,letterpaper]{article}

\usepackage[pagenumbers]{cvpr} 

\usepackage{graphicx}
\usepackage{amsmath}
\usepackage{amssymb}
\usepackage{booktabs}
\usepackage{multirow}
\usepackage[table]{xcolor}   
\usepackage{algorithm}
\usepackage{algpseudocode}
\usepackage{etoolbox}

\makeatletter
\newcommand*{\algrule}[1][\algorithmicindent]{%
	\makebox[#1][l]{%
		\hspace*{.2em}
		\vrule height .75\baselineskip depth .25\baselineskip
	}
}

\newcount\ALG@printindent@tempcnta
\def\ALG@printindent{%
	\ifnum \theALG@nested>0
	\ifx\ALG@text\ALG@x@notext
	\else
	\unskip
	\ALG@printindent@tempcnta=1
	\loop
	\algrule[\csname ALG@ind@\the\ALG@printindent@tempcnta\endcsname]%
	\advance \ALG@printindent@tempcnta 1
	\ifnum \ALG@printindent@tempcnta<\numexpr\theALG@nested+1\relax
	\repeat
	\fi
	\fi
}
\patchcmd{\ALG@doentity}{\noindent\hskip\ALG@tlm}{\ALG@printindent}{}{\errmessage{failed to patch}}
\patchcmd{\ALG@doentity}{\item[]\nointerlineskip}{}{}{} 
\makeatother

\usepackage[pagebackref,breaklinks,colorlinks,bookmarks=false]{hyperref}

\usepackage[capitalize]{cleveref}
\crefname{section}{Sec.}{Secs.}
\Crefname{section}{Section}{Sections}
\Crefname{table}{Table}{Tables}
\crefname{table}{Tab.}{Tabs.}

\def\cvprPaperID{4019} 
\def\confName{CVPR}
\def\confYear{2022}

\begin{document}

\title{Self-Supervised Predictive Learning: A Negative-Free Method for\\Sound Source Localization in Visual Scenes}

\author{Zengjie Song$^{1}$
	\quad Yuxi Wang$^{1,3}$
	\quad Junsong Fan$^{1,2}$
	\quad Tieniu Tan$^{1,2}$
	\quad Zhaoxiang Zhang$^{1,2,3}$\thanks{Corresponding author.}\\%
	$^{1}$Center for Research on Intelligent Perception and Computing, NLPR, CASIA\\%
	$^{2}$University of Chinese Academy of Sciences (UCAS)\\%
	$^{3}$Centre for Artificial Intelligence and Robotics, HKISI\_CAS\\%
	{\tt\small \{zengjie.song, wangyuxi2016, fanjunsong2016, zhaoxiang.zhang\}@ia.ac.cn, tnt@nlpr.ia.ac.cn}%
}

\maketitle

\begin{abstract}
	Sound source localization in visual scenes aims to localize objects emitting the sound in a given image. Recent works showing impressive localization performance typically rely on the contrastive learning framework. However, the random sampling of negatives, as commonly adopted in these methods, can result in misalignment between audio and visual features and thus inducing ambiguity in localization. In this paper, instead of following previous literature, we propose Self-Supervised Predictive Learning (SSPL), a negative-free method for sound localization via explicit positive mining. Specifically, we first devise a three-stream network to elegantly associate sound source with two augmented views of one corresponding video frame, leading to semantically coherent similarities between audio and visual features. Second, we introduce a novel predictive coding module for audio-visual feature alignment. Such a module assists SSPL to focus on target objects in a progressive manner and effectively lowers the positive-pair learning difficulty. Experiments show surprising results that SSPL outperforms the state-of-the-art approach on two standard sound localization benchmarks. In particular, SSPL achieves significant improvements of 8.6\% cIoU and 3.4\% AUC on SoundNet-Flickr compared to the previous best. Code is available at: \href{https://github.com/zjsong/SSPL}{https://github.com/zjsong/SSPL}.%
\end{abstract}

\section{Introduction}\label{sec:intro}
\begin{figure}
	\centering
	\includegraphics[width=\linewidth]{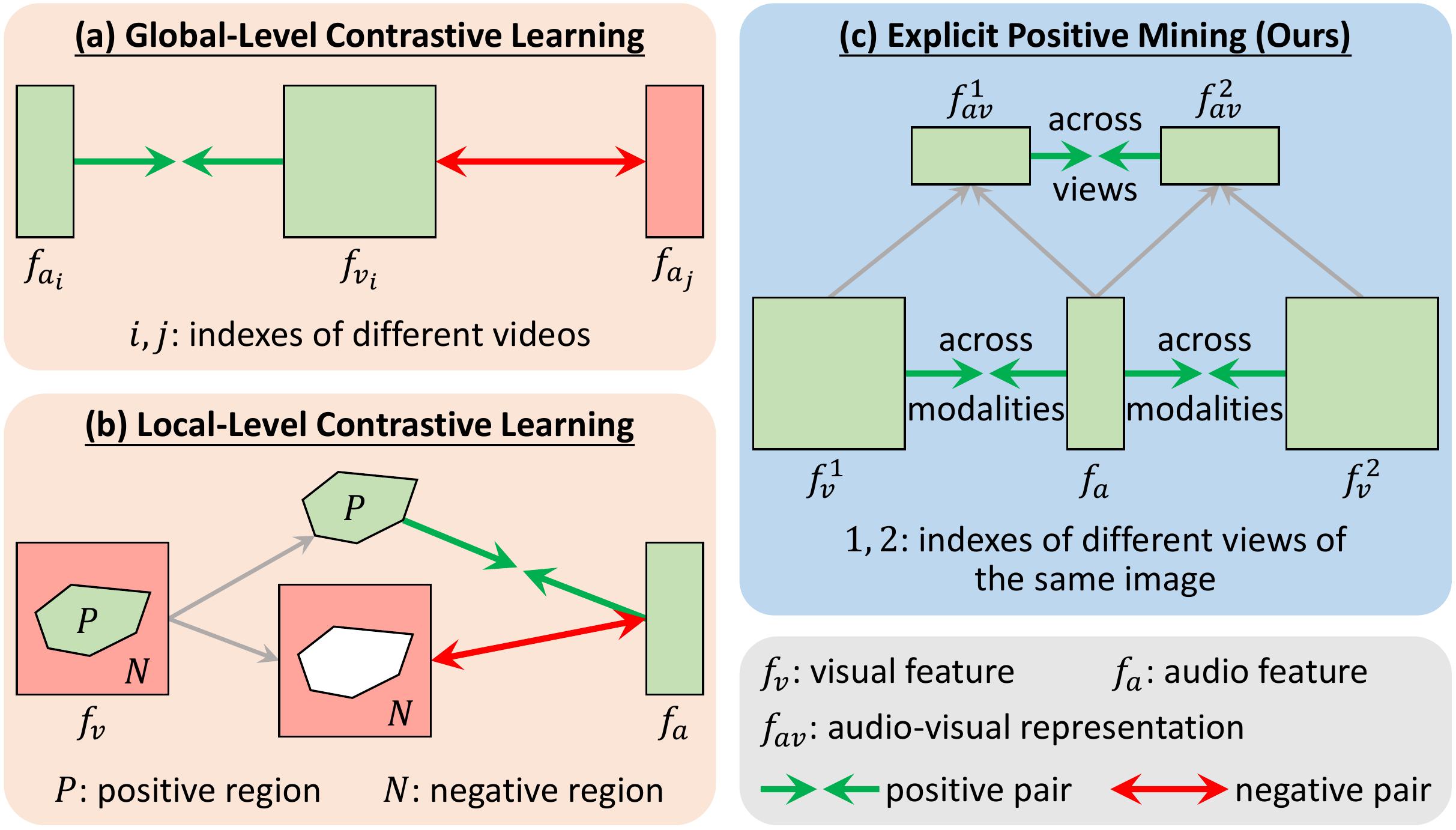}%
	\vspace{-2mm}
	\caption{\textbf{Three methods to build audio-visual correspondence for sound localization.} Previous contrastive learning based methods have to construct \emph{negative} pairs (a) at the global feature level or (b) at the local feature level. (c) Our method, by contrast, explores the coherent similarities between audio and visual features by only explicit \emph{positive} mining.}%
	\label{fig:comp_methods}
	\vspace{-3mm}
\end{figure}

When strolling in a park brimming with life, you notice that the bird sitting on a twig is chirping; the puppy on the road ahead gives a little bark; and after a while an acquaintance may walk by and say hello to you friendly. Despite a short notice, humans own the excellent ability to associate the sounds they hear with the corresponding visual perception, and thus can localize and distinguish different sounding objects from one another. 

To mimic humans' such ability, in this work, we pay attention to the task of sound source localization in visual scenes, where the goal is to localize regions of the visual landscape that correlate highly with the audio cues. While handling this task is a long-standing challenge \cite{Hershey99,Fisher00}, remarkable breakthroughs have been made until recent progresses on self-supervised audio-visual learning \cite{Owens18,Senocak18,Hu19,Qian20,Afouras20,Hu20b}. These methods leverage the free supervision rooted in videos, \eg, the natural correspondence and/or temporal synchronization between audio and visual sources, to guide multi-modal feature extraction and alignment; then the similarity map between audio and visual features is usually employed to localize sounding objects. Among them, contrastive learning has particularly achieved impressive performance on this task \cite{Senocak18,Qian20,Afouras20,Chen21c,Tian21}.  

Existing contrastive learning methods in this line of work can be cast into two categories: the first one is global-level contrastive learning (GLCL, \Cref{fig:comp_methods}a) \cite{Senocak18,Morgado20,Qian20,Afouras20}, which commonly attracts audio and visual features extracted from the same video and repulses features from different videos; the other one is local-level contrastive learning (LLCL, \Cref{fig:comp_methods}b) \cite{Tian21,Chen21c,Lin21}, which further compares audio feature with different visual feature components, even though they have correspondence at the video level. Generally, to perform contrastive learning, these methods randomly sample sounds to form negative pairs with the given video frame. However, this randomness can produce false negatives by sampling sounds that actually belong to the same category as the positive one, and thus hampering the model to align audio and visual features in semantic level. The misalignment as a result induces the learning process to build inaccurate audio-visual correspondence for localization. 

\begin{figure}
	\centering
	\includegraphics[width=\linewidth]{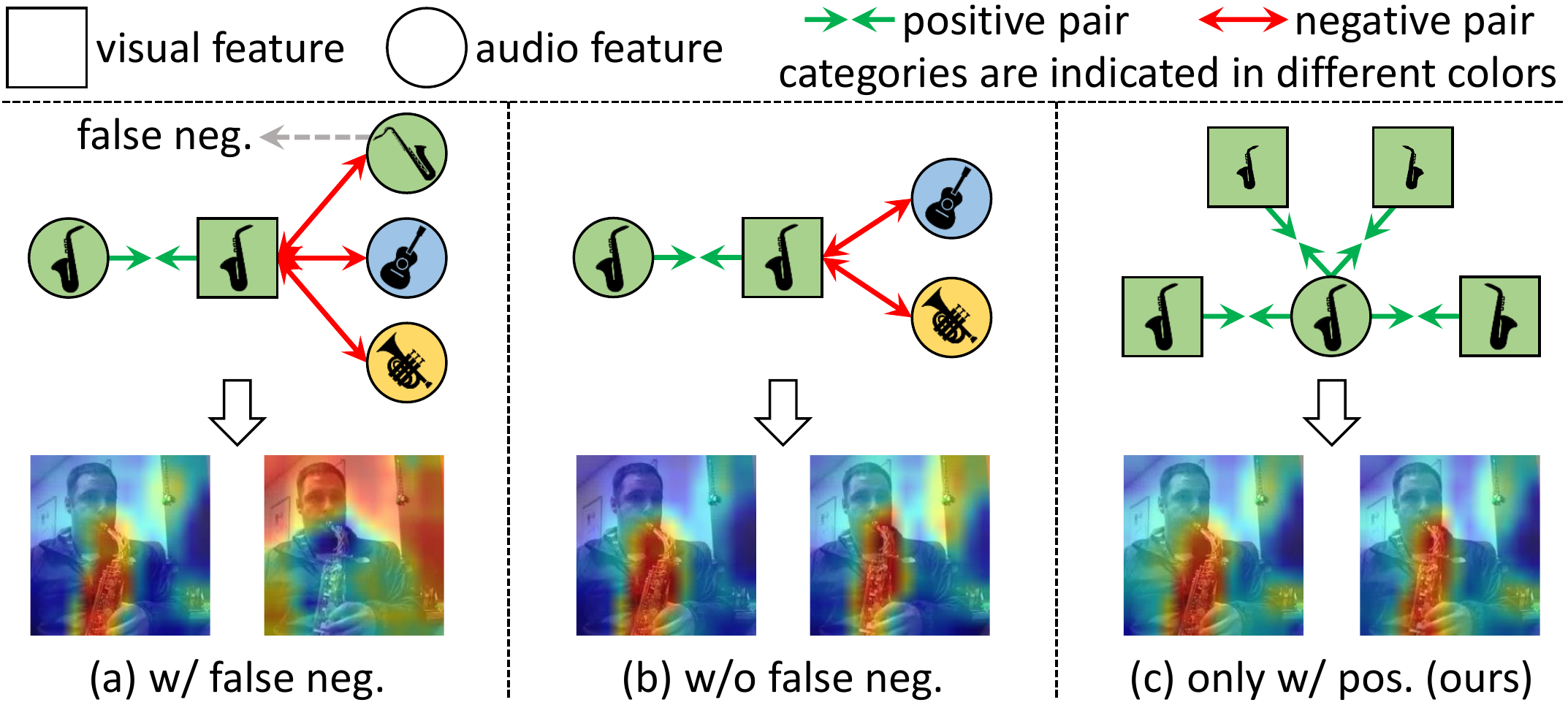}%
	\vspace{-2mm}
	\caption{\textbf{Effect of false negatives on sound localization.} (a) Ambiguity in localization is observed when learning with false negatives that have the same class label as the positive one. (b) Consistent localization can be obtained without sampling false negatives, but requiring \emph{class label} as guidance. (c) Our method mitigates this problem via self-supervised positive mining. Experiments are performed on MUSIC \cite{Zhao18} where labels are available.}%
	\label{fig:effect_negatives}
	\vspace{-3mm}
\end{figure}

We conduct a pilot experiment to illustrate the effect of such false negatives in \Cref{fig:effect_negatives}. Given image and sound from the same video (\eg, saxophone playing) as positive pair, other videos' sounds, holding the same category as the positive one, are allowed to construct negative pairs in \Cref{fig:effect_negatives}a, while not allowed in \Cref{fig:effect_negatives}b. We keep the remaining training settings same for these two cases. During testing, consequently, the former case generates ambiguous localization on sounding objects (\ie, saxophone here) and the later one not. Based on this observation, we take a step back and ask the questions: Do we really need negatives to develop self-supervised sound localization methods? Can the image-audio positive pair alone be used to achieve the same goal?

To answer these questions, we propose Self-Supervised Predictive Learning (SSPL), a negative-free approach for sound source localization through explicit audio-visual positive mining. The predictive learning is embodied from two perspectives (\Cref{fig:comp_methods}c): predicting across different visual views and predicting across audio and visual modalities. Given an image-audio pair from one video, the former perspective hypothesizes that if two different visual views of one video frame contain the same sounding objects, they should share a consistent correspondence with the given sound source. We achieve this consistency by the mutual prediction of two audio-visual representations. For the later perspective, we devise a predictive coding module (PCM) that uses visual features to iteratively predict audio ones, providing a coarse-to-fine way to automatically align features. To the best of our knowledge, this is the first attempt to apply the self-supervised negative-free method to the audio-visual task of sound localization.

Our main contributions can be summarized as follows:
\begin{itemize}
	\item We propose a novel negative-free method to extend a self-supervised learning framework to the audio-visual data domain for sound localization, and show how it can effectively address the false negative sampling problem.
	
	\item We propose the predictive coding module for feature alignment, which enables the model to progressively attend to relevant visual features while ignoring information irrelevant to audio cues, boosting sound localization significantly.
	
	\item Comprehensive experiments demonstrate the effectiveness of the proposed approach, which achieves localization performance superior to the state-of-the-art on SoundNet-Flickr and VGG-Sound Source. 
\end{itemize}

\section{Related Work}\label{sec:related_work}
\noindent\textbf{Self-Supervised Visual Representation Learning.} Self-supervised learning (SSL) has achieved remarkable breakthroughs on large computer vision benchmarks. Most of the current SSL methods \cite{Henaff20,Chen20a,Chen20b,Chen20c,He20,Tian20,Feichtenhofer21,Chen21d,Qian21} resort to the design of contrastive learning strategy \cite{Oord18}. These methods, at their core, transform one image into multiple views, and repulse different images (negatives) meanwhile attracting the same image’s different views (positives). Recently, several efforts have been made to further relieve the requirement of negatives and simplify the SSL framework beyond conventional contrastive learning, including BarlowTwins \cite{Zbontar21}, W-MSE \cite{Ermolov21}, BYOL \cite{Grill20}, and SimSiam \cite{Chen21a}. In SimSiam \cite{Chen21a}, researchers investigate the importance of simple Siamese architecture for unsupervised representation learning, and empirically show that the stop-gradient operation is critical for the network to prevent collapse, even without using momentum encoder \cite{He20} and large batches \cite{Chen20a}. These advances in image representation learning provide insights for our work to develop effective audio-visual SSL method. 

\begin{figure*}[t]
	\centering
	\includegraphics[width=\linewidth]{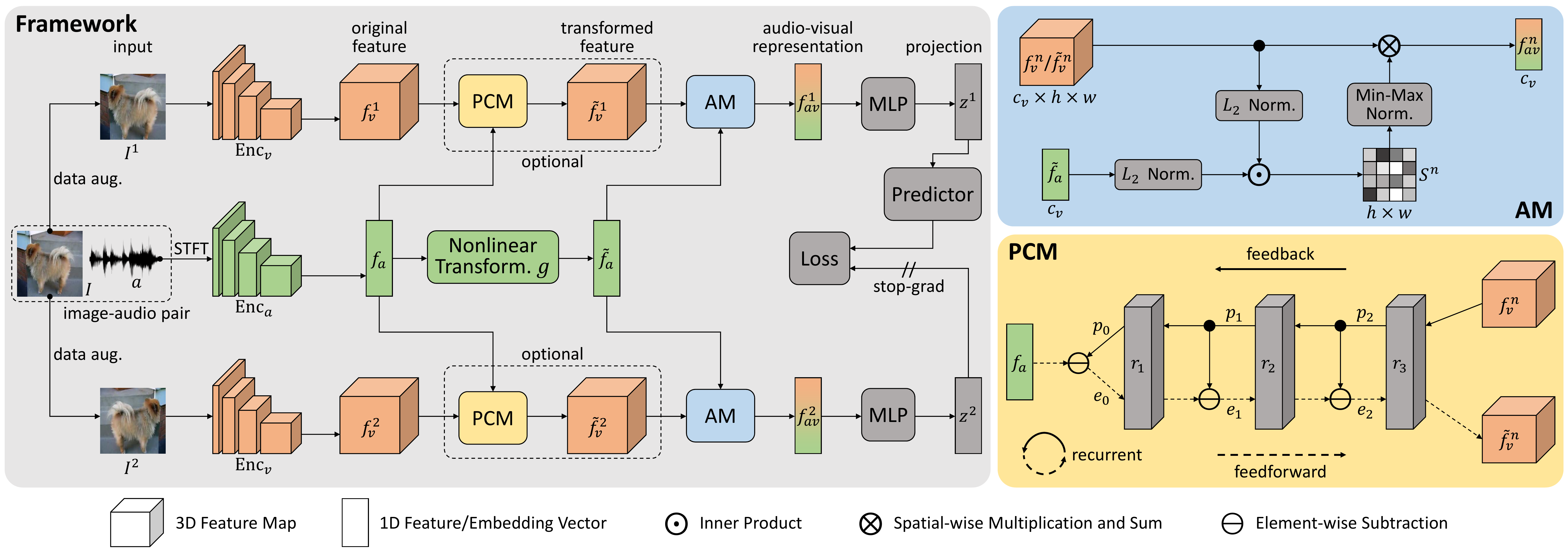}%
	\vspace{-1mm}
	\caption{\textbf{Left:} framework of our SSPL method. \textbf{Top right:} attention module (AM) to compute audio-visual representation. \textbf{Bottom right:} predictive coding module (PCM) to align audio and visual features (for simplicity we only show a 3-layer version). In the framework, $\text{Enc}_{a}$ of the middle processing stream derives the discriminative audio feature of audio signal $a$, while $\text{Enc}_{v}$ extracts visual features of two augmented images $I^{1}$ and $I^{2}$. Note that $\text{Enc}_{v}$ shares weights between top and bottom streams, and a similar setting holds for PCM and MLP. During inference, the similarity map $S^{n}$, generated in AM, is resized to image scale and treated as the sound localization map.}%
	\label{fig:framework_sspl}
\end{figure*}

\noindent\textbf{Audio-Visual Representation Learning.} The vision and sound are usually two co-occurring modalities, which can naturally be used to derive supervisions for audio-visual learning \cite{Aytar16,Owens16,Korbar18,Chen20e,Su21}. In \cite{Aytar16}, for instance, the visual features extracted from pretrained teacher networks act to guide the student network to learn more discriminative sound representation, and vice versa in \cite{Owens16}. Korbar \etal \cite{Korbar18} and Owens and Efros \cite{Owens18} leverage the synchronization between audio and visual streams to build negative samples and contrastive losses, obtaining versatile multisensory features, respectively. Several works also explore the audio-visual correspondence by feature clustering \cite{Hu19,Hu20a,Alwassel20}. In general, these methods focus on learning task-agnostic representations, which work well on classification-related down-stream tasks, such as action/scene recognition \cite{Korbar18,Owens18,Morgado20,Alwassel20,Chen21b}, audio event classification \cite{Arandjelovic17,Korbar18,Hu19,Alwassel20,Morgado21}, video retrieval \cite{Chen21b}, \emph{etc}. However, they are not customized for sound source localization, and as a result only achieve limited performance on this task \cite{Owens18,Hu19,Hu20a}.

\noindent\textbf{Sound Localization in Visual Scenes.} Early works to solve this task mainly rely on statistical modeling of the cross-modal relationship by using, for example, mutual information \cite{Hershey99,Fisher00} and canonical correlation analysis \cite{Kidron05,Izadinia12}. However, these methods as shallow models only show advantages in simple audio-visual scenarios. By digging into the correspondence between deep audio and visual features, recent deep learning methods give promising solutions to this problem \cite{Senocak18,Hu19,Gan19,Qian20,Afouras20,Chen21c,Tian21,Valverde21}. For instance, Senocak \etal \cite{Senocak18} employ a two-stream framework and an attention mechanism to compute sound localization map. Qian \etal \cite{Qian20} achieve the same goal by using the class activation map (CAM) derived from a weakly-supervised approach. In \cite{Hu19,Hu20a}, audio and visual features are clustered, respectively, and the assignment weights based on the distance between features and cluster centers are adopted to localize sounding objects. In addition to viewing image and sound extracted from different videos as negative pair, Chen \etal \cite{Chen21c} and Lin \etal \cite{Lin21} propose to mine hard negatives within an image-audio pair, \ie, background regions that correlate lowly with the given sound are treated as extra hard negatives. Different from these negative-based works, we handle the same task by explicit positive mining, providing an effective alternative for sound localization.

\section{Method}
\Cref{fig:framework_sspl} depicts the overall framework of our SSPL, which is a three-stream network, making a big difference with widely-used two-stream ones. The top and bottom streams first serve to extract deep visual features from different views of the same image. Then, they employ PCM and AM to integrate visual and audio features, where the discriminative audio feature is derived from the middle stream subnetwork. Subsequently, two audio-visual representations are enforced to be similar by self-supervised predicting with each other. The sound localization map is a natural consequence of representation learning and is generated in the AM. Note that the vanilla SSPL without PCM focuses on exploring audio-visual correspondence across different image views (\cref{sec:sspl_across_views}), while the PCM component excels at aligning features across modalities (\cref{sec:sspl_across_modalities}), and thus boosting localization performance further. We elaborate on and formulate each part in the following.

\subsection{Unimodal Features of Audio and Vision}
Let $I \in \mathbb{R}^{3\times H_{v}\times W_{v}}$ and $a\in\mathbb{R}^{H_{a}\times W_{a}}$ denote a video frame and a corresponding audio signal from the same video clip, respectively. Here the raw 1D audio waveform has been converted into the 2D spectrogram by Short-Time Fourier Transform (STFT), and therefore we use 2D CNNs to extract deep semantic features of audio modality like vision. In practice, we employ the off-the-shelf VGG16 \cite{Simonyan15} for frame processing ($\text{Enc}_{v}$) and the VGGish network \cite{Hershey17} for spectrogram analysis ($\text{Enc}_{a}$), similar to \cite{Hu20a}. The output feature map of the final convolution layer of VGG16 is treated as the original visual feature $f_{v}$. We use layers before the final post-processing stage of VGGish to produce a high-level embedding as the original audio feature $f_{a}$. These feature extraction processes are formulated as:
\begin{align}
	f_{v} &= \text{Enc}_{v}(I), \quad f_{v}\in \mathbb{R}^{c_{v}\times h\times w}, \label{eq:orig_visual_feat}\\
	f_{a} &= \text{Enc}_{a}(a), \quad f_{a}\in \mathbb{R}^{c_{a}}. \label{eq:orig_audio_feat}
\end{align}

\subsection{Predictive Learning across Visual Views}\label{sec:sspl_across_views}
This section details the vanilla SSPL, which contains the attention module to compute audio-visual representation from previously extracted features, and the self-supervised learning to guide model training via cross-view representation prediction.

Let $I^{1}$ and $I^{2}$ denote two randomly augmented views of the given image $I$. The two views are respectively fed into the visual CNN, $\text{Enc}_{v}$, to obtain spatial feature maps $f_{v}^{1}$ and $f_{v}^{2}$ as in \cref{eq:orig_visual_feat}. Besides, an audio feature vector $f_{a}$ is derived from the audio signal $a$ using \cref{eq:orig_audio_feat}. For simplicity, we use $n\in\{1,2\}$ to index different visual views.

\noindent\textbf{Attention Module (AM).} We adopt the normalized inner product (or cosine similarity) to measure the similarity between audio and visual features, as suggested by \cite{Senocak18,Chen21c}. Considering that $f_{a}$ and $f_{v}^{n}$ are from two heterogeneous modalities, we first transform $f_{a}$ to be comparable with the visual feature via a nonlinear transformation, $\tilde{f}_{a}=g(f_{a})\in\mathbb{R}^{c_{v}}$, and then perform the similarity measurement. Formally, for the spatial location $(i,j)$ in visual feature map $f_{v}^{n}$, a similarity value is computed as follows:
\begin{equation}\label{eq:cosine_similarity}
	S^{n}(i,j) = \frac{\langle \tilde{f}_{a}, f_{v}^{n}(\cdot,i,j) \rangle}{\|\tilde{f}_{a}\|_{2}\|f_{v}^{n}(\cdot,i,j)\|_{2}}, \quad (i,j) \in [h]\times[w],
\end{equation}
where $f_{v}^{n}(\cdot,i,j)\in \mathbb{R}^{c_{v}}$. 

The similarity map $S\in\mathbb{R}^{h\times w}$ plays two important roles in our method. On the one hand, it indicates the degree of correlation between each image location (after resized to image scale) and the given audio cues, and thus can serve as the sound localization map. On the other hand, it acts as an attention mechanism to weigh the original visual feature, resulting in the following audio-visual representation:
\begin{align}
	f_{av}^{n}(k) &= \sum_{i,j}\tilde{S}^{n}(i,j)f_{v}^{n}(k,i,j), \quad k\in\{1,\dots,c_{v}\}, \label{eq:av_repres}\\
	\tilde{S}^{n} &= \frac{S^{n}-\min(S^{n})}{\max(S^{n})-\min(S^{n})}. \label{eq:norm_sim_map}
\end{align}
Here we scale the similarity values to $[0,1]$ by min-max normalization \cite{Lin21}. This operation makes different feature elements more distinguishable, and performs better compared with the sigmoid and softmax scaling functions \cite{Senocak18,Qian20} (see \Cref{tab:scale_method} for empirical comparisons). Since the $f_{av}^{n}\in \mathbb{R}^{c_{v}}$ selects and integrates visual features that are more related to audio ones, we treat it as a multi-modal representation to advance subsequent learning. 

\noindent\textbf{Self-Supervised Learning.} 
The learning procedure aims to make the two audio-visual representations similar. Our hypothesis is that two visual scenes containing the same sounding objects should consistently correspond to the same audio cues in semantic level. We follow SimSiam \cite{Chen21a} to achieve this goal in the audio-visual setting.

Formally, we feed $f_{av}^{n}$ into a MLP head to obtain the projection of corresponding view, $z^{n}=\text{MLP}(f_{av}^{n})$. Then a predictor head, denoted as Pred, takes as input $z^{1}$ to predict $z^{2}$ by minimizing the negative cosine similarity (NCS):
\begin{equation}\label{eq:neg_cos_sim}
	\mathcal{L}_{NCS}(z^{1}, z^{2}) = -\frac{\langle \text{Pred}(z^{1}), z^{2} \rangle}{\|\text{Pred}(z^{1})\|_{2}\|z^{2}\|_{2}}.
\end{equation}
To symmetrize the above loss, we also feed $z^{2}$ into Pred to estimate $z^{1}$, leading to another loss term $\mathcal{L}_{NCS}(z^{2}, z^{1})$. The total loss is therefore defined as:
\begin{equation}\label{eq:simsiam_loss0}
	\mathcal{L}_{SSPL} = \frac{1}{2}\mathcal{L}_{NCS}(z^{1}, z^{2}) + \frac{1}{2}\mathcal{L}_{NCS}(z^{2}, z^{1}).
\end{equation} 
However, as discussed in \cite{Chen21a}, directly minimizing the loss in \cref{eq:simsiam_loss0} could easily induce representation collapse. To overcome this problem the stop-gradient (SG) operation is employed. That is, \cref{eq:neg_cos_sim} is modified as $\mathcal{L}_{NCS}(z^{1}, \text{SG}(z^{2}))$, where $z^{2}$ is viewed as a constant such that branch on $I^{2}$ receives no gradient from $z^{2}$ through this loss term. Similarly we have $\mathcal{L}_{NCS}(z^{2}, \text{SG}(z^{1}))$, and the form in \cref{eq:simsiam_loss0} is implemented as:
\begin{equation}\label{eq:simsiam_loss}
	\mathcal{L}_{SSPL} = \frac{1}{2}\mathcal{L}_{NCS}(z^{1}, \text{SG}(z^{2})) + \frac{1}{2}\mathcal{L}_{NCS}(z^{2}, \text{SG}(z^{1})).
\end{equation}

Note that we follow the SimSiam framework for its simplicity in the use of only positive pairs without representation collapse. However, our predictive learning strategy can be combined with other self-supervised learning methods, such as BYOL \cite{Grill20}, W-MSE \cite{Ermolov21}, and BarlowTwins \cite{Zbontar21}. We leave the potential extension for future works.

\subsection{Predictive Learning across Modalities}\label{sec:sspl_across_modalities}
In this section, we propose the PCM for audio and visual feature alignment, and continuously improving the localization performance of SSPL. The key idea inherits the spirit of predictive coding (PC) in neuroscience \cite{Rao99,Wen18,Song18}, which simulates the mechanism of information processing in visual cortex. In brief, PC uses feedback connections from a higher-level area to a lower-level one to convey predictions of lower-level neural activities; it employs feedforward connections to carry the errors between the actual activities and the predictions; and the brain dynamically updates representations so as to progressively reduce the prediction errors. In our PCM (\Cref{fig:framework_sspl}), we treat the visual feature as a type of prior knowledge to predict the audio feature in an \emph{iterative} manner. In the following, at the heart of PCM, we give the representation update rules of feedback and forward processes, respectively. The detailed derivations can be found in supplement.

Denote by $r_{l}(t), l\in\{1,\dots,L\}, t\in\{0,\dots,T\}$ the representation of the $l$-th layer of PCM network at time step $t$, and by $W_{l,l-1}$ the feedback connection weights from layer $l$ to layer $l-1$ (and vice versa for $W_{l-1,l}$).

The \textbf{\emph{feedback process}} updates representations through a mechanism of layer-wise prediction generation. Concretely, at $l$-th layer the prediction, $p_{l}$, of representation, $r_{l}$, is first derived using the above layer's representation, $r_{l+1}$. Then $r_{l}$ is updated with its previous state and the prediction, \ie, at time step $t$ we have:
\begin{align}
	p_{l}(t)   &= (W_{l+1,l})^{T}r_{l+1}(t), \label{eq:feedback_pred}\\
	r_{l}(t) &\gets \phi((1-b_{l})r_{l}(t-1) + b_{l}p_{l}(t)), \label{eq:feedback_repres} 
\end{align}
where $\phi$ is a nonlinear activation function and $b_{l}$ serves as a positive scalar to balance two terms. The above update rules are executed from top layer $L$ to bottom layer $1$ in sequence, and by setting $p_{L}(t)\equiv f_{v}^{n}$, we in fact achieve further feature extraction from visual source. 

In \textbf{\emph{feedforward process}}, representations are again modulated based on prediction errors emerged at each layer. Specifically, the representation $r_{l-1}$ and its prediction $p_{l-1}$ are often unequal, resulting in a prediction error $e_{l-1}$. The error signal contains unpredictable components of $r_{l-1}$, and is forwarded to higher level to correct the representation $r_{l}$. This leads to complementary update rules:
\begin{align}
	e_{l-1}(t) &= r_{l-1}(t) - p_{l-1}(t), \label{eq:feedforward_error}\\
	r_{l}(t) &\gets \phi(r_{l}(t) + a_{l}(W_{l-1,l})^{T}e_{l-1}(t)), \label{eq:feedforward_repres}
\end{align}
where $r_{0}(t)\equiv f_{a}$ is the original audio feature, $p_{0}(t)=\phi((W_{1,0})^{T}r_{1}(t))$ refers to the prediction of $f_{a}$, and $a_{l}$ denotes a trade-off scalar like $b_{l}$.

PCM conducts the two distinct processes alternatively while all layers' representations are progressively refined so as to reduce the prediction error. Subsequently, we use a $1\times1$ convolution to transform the top layer representation at last time step, $r_{L}(T)$, to a new visual feature, $\tilde{f}_{v}^{n}$, with the same dimension of $f_{v}^{n}$. Consequently the vanilla SSPL method in \cref{sec:sspl_across_views} can be enhanced by feeding $\tilde{f}_{v}^{n}$, instead of $f_{v}^{n}$, into the AM to compute audio-visual representation (\ie, \cref{eq:cosine_similarity,eq:av_repres}).

\section{Experiments}
\subsection{Datasets and Evaluation Metrics}
\noindent\textbf{SoundNet-Flickr \cite{Aytar16}.} This dataset consists of more than 2 million videos from Flickr. We use a 3s audio clip around the middle frame of the whole audio, and the accompanied video frame to form an image-audio pair. Following \cite{Senocak18,Chen21c}, we train models with two random subsets of 10k and 144k image-audio pairs, respectively, and perform evaluation on the 250 annotated pairs provided by \cite{Senocak18}. Note that the location of the sound source in each test frame is given by 3 separate bounding boxes, each of which is obtained by a different annotator.

\noindent\textbf{VGG-Sound \cite{Chen20d} and VGG-Sound Source \cite{Chen21c}.} VGG-Sound dataset contains over 200k video clips that are divided into 300 sound categories. Similar to \cite{Chen21c}, we conduct training with 10k and 144k image-audio pairs randomly sampled from this dataset, respectively. For fair comparisons with recent works \cite{Senocak18,Afouras20,Chen21c}, we evaluate models on the VGG-Sound Source (VGG-SS) benchmark with 5k annotated image-audio pairs collected by \cite{Chen21c}. Compared with SoundNet-Flickr benchmark that spans about 50 sounding object classes, VGG-SS has 220 classes and thus providing a more challenging scenario for sound localization task.

We focus on and reimplement two related methods, Attention \cite{Senocak18} and HardWay \cite{Chen21c} (SOTA on this task), which could be representatives of GLCL- and LLCL-based approaches, respectively. We denote our method without using PCM by SSPL (w/o PCM), and the version equipped with PCM by SSPL (w/ PCM). Additionally, we employ consensus Intersection over Union (cIoU) and Area Under Curve (AUC) as evaluation metrics, and report cIoU scores with threshold 0.5 in experiments, same as \cite{Senocak18,Qian20,Chen21c,Lin21}.

\subsection{Implementation Details}
We use VGG16 \cite{Simonyan15} pretrained on ImageNet \cite{Deng09} and VGGish \cite{Hershey17} pretrained on AudioSet \cite{Gemmeke17} as visual and audio feature extractors, respectively. The visual input is an image of size $256\times256\times3$, on which we perform the data augmentation pipeline: random cropping with $224\times224$ resizing and random horizontal flip. The raw 3s audio signal is re-sampled at 16kHz and further transformed into $96\times64$ log-mel spectrograms as audio input, and the audio output feature $f_{a}$ is a 128D vector. The nonlinear audio feature transformation function $g(\cdot)$ is instantiated with a simple two-layer network as in \cite{Senocak18}: \texttt{FC}(512)-\texttt{ReLU}-\texttt{FC}(512). We closely follow SimSiam \cite{Chen21a} to set the projection and prediction MLPs. For PCM, we mainly adopt \texttt{Conv}-\texttt{MaxPool}-\texttt{GELU} layers in the feedback pathway, and \texttt{Upsample}-\texttt{DeConv}-\texttt{GELU} layers in the feedforward counterpart. The weights of two feature extractors are kept frozen during training, and we optimize the rest of the model with AdamW \cite{Loshchilov19}. We utilize the early stopping strategy to avoid overfitting in all cases. More setting details (\eg, learning rate and batch size) are in supplement.

\subsection{Comparisons with State-of-the-art Methods}
\begin{table}
	\tabcolsep=8.3pt
	\footnotesize
	\centering
	\begin{tabular}{llcc}
		\toprule
		Method    							    				&Training set  		&cIoU $\uparrow$       &AUC $\uparrow$ \\
		\midrule\midrule
		Attention \cite{Senocak18}$_{\text{CVPR}18}^{\dagger}$  &Flickr10k     		&0.442                 &0.461 \\
		DMC \cite{Hu19}$_{\text{CVPR}19}$						&Flickr10k          &0.414				   &0.450 \\
		CAVL \cite{Hu20a}$_{\text{arXiv}20}$					&Flickr10k          &0.500				   &0.492 \\
		MSSL \cite{Qian20}$_{\text{ECCV}20}$      			 	&Flickr10k     		&0.522    			   &0.496 \\
		AVObject \cite{Afouras20}$_{\text{ECCV}20}$  			&Flickr10k     		&0.546 			       &0.504 \\
		DSOL \cite{Hu20b}$_{\text{NeurIPS}20}$					&Flickr10k			&0.566 				   &0.515 \\
		HardWay \cite{Chen21c}$_{\text{CVPR}21}^{\dagger}$   	&Flickr10k     		&0.615 				   &0.535 \\
		ICL \cite{Lin21}$_{\text{CVPRW}21}$					 	&Flickr10k			&0.710				   &0.580 \\
		\rowcolor{lightgray!25}
		SSPL (w/o PCM)   					 					&Flickr10k     		&0.671				   &0.556 \\
		\rowcolor{lightgray!25}
		SSPL (w/ PCM)    					 					&Flickr10k     		&\textbf{0.743}		   &\textbf{0.587} \\
		\midrule
		Attention \cite{Senocak18}$_{\text{CVPR}18}$  		 	&Flickr144k	   		&0.660	   			   &0.558 \\
		DMC \cite{Hu19}$_{\text{CVPR}19}$						&Flickr144k	   		&0.671 				   &0.568 \\
		HardWay \cite{Chen21c}$_{\text{CVPR}21}^{\dagger}$	 	&Flickr144k	   		&0.699				   &0.590 \\
		\rowcolor{lightgray!25}
		SSPL (w/o PCM)   					 					&Flickr144k    		&0.699				   &0.580 \\
		\rowcolor{lightgray!25}
		SSPL (w/ PCM)    					 					&Flickr144k	   		&\textbf{0.759}		   &\textbf{0.610} \\
		\midrule
		Attention \cite{Senocak18}$_{\text{CVPR}18}^{*}$	 	&VGG-Sound10k       &0.522                 &0.502 \\
		HardWay \cite{Chen21c}$_{\text{CVPR}21}^{\dagger}$   	&VGG-Sound10k  		&0.647                 &0.560 \\
		\rowcolor{lightgray!25}
		SSPL (w/o PCM)   					 					&VGG-Sound10k  		&0.699				   &0.572 \\
		\rowcolor{lightgray!25}
		SSPL (w/ PCM)    					 					&VGG-Sound10k  		&\textbf{0.763}		   &\textbf{0.591} \\
		\midrule
		HardWay \cite{Chen21c}$_{\text{CVPR}21}^{\dagger}$	 	&VGG-Sound144k   	&0.723	   			   &\textbf{0.605} \\
		HardWay \cite{Chen21c}$_{\text{CVPR}21}$      		 	&VGG-Sound Full  	&0.735				   &0.590 \\
		\rowcolor{lightgray!25}
		SSPL (w/o PCM)   					 					&VGG-Sound144k    	&0.739	               &0.602 \\
		\rowcolor{lightgray!25}
		SSPL (w/ PCM)    					 					&VGG-Sound144k  	&\textbf{0.767}        &\textbf{0.605} \\
		\bottomrule
	\end{tabular}%
	\vspace{-2.5mm}
	\caption{\textbf{Quantitative localization results on SoundNet-Flickr test set.} ``$*$'' denotes our reproduction, and ``$\dagger$'' indicates \emph{improved} reproduction \vs original papers (see supplement).}%
	\label{tab:compare_flickr}
	\vspace{-3mm}
\end{table}

We first compare SSPL with recent methods on the SoundNet-Flickr test set in \Cref{tab:compare_flickr}. We observe that when trained by 10k Flickr samples, the vanilla SSPL (w/o PCM) performs favorably against the two competing methods, HardWay \cite{Chen21c} and ICL \cite{Lin21}, while the enhanced SSPL (w/ PCM) outperforms the previous best \cite{Lin21} by a large margin (0.710 \vs 0.743, around 5\% improvement). In the Flickr144k training case, SSPL (w/ PCM) increases performance by 8.6\% cIoU and 3.4\% AUC compared to HardWay, establishing a new state-of-the-art on this benchmark. These results demonstrate that SSPL without relying on negatives is feasible and effective for sound localization. Following \cite{Chen21c}, we also train on VGG-Sound using respective 10k and 144k data pairs, which enables SSPLs to achieve the top two localization performance in both settings. As discussed in \cite{Chen21c}, the sounding objects are often visible in video clips from VGG-Sound, revealing that our method can benefit from the improved data quality. What's more, the performance of SSPL is significantly boosted by PCM, especially in the 10k's setting (11\% improvement for Flickr10k and 9\% for VGG-Sound10k). This illustrates the advantage of PCM for facilitating sound localization.

We further evaluate SSPL on the newly released VGG-SS benchmark and report results in \Cref{tab:compare_vggss}. Because in this challenging benchmark the sounding object categories are more diverse and the number of test samples is greater than those of SoundNet-Flickr \cite{Chen21c}, the performance of all methods drops severely compared with the results in \Cref{tab:compare_flickr}. While SSPL (w/o PCM) still outperforms Attention by a large margin, it does not overtake HardWay. We attribute this to the limitation of vanilla SSPL on dealing with background noise (see \cref{sec:qualit_analysis} for an empirical comparison). However, by combining with feature alignment module, SSPL (w/ PCM) yields performance better than the state-of-the-art HardWay, especially by a substantial gap in the 10k's scenario (0.277 \vs 0.314, over 13\% gain). This verifies the superiority of the enhanced SSPL.
\begin{table}
	\tabcolsep=8.2pt
	\footnotesize
	\centering
	\begin{tabular}{llcc}
		\toprule
		Method    					        				&Training set       &cIoU $\uparrow$       &AUC $\uparrow$ \\
		\midrule\midrule
		Attention \cite{Senocak18}$_{\text{CVPR}18}^{*}$    &VGG-Sound10k   	&0.160                 &0.283 \\
		HardWay \cite{Chen21c}$_{\text{CVPR}21}^{*}$   	    &VGG-Sound10k   	&0.277 				   &0.349 \\
		\rowcolor{lightgray!25}
		SSPL (w/o PCM)   									&VGG-Sound10k   	&0.253				   &0.335 \\
		\rowcolor{lightgray!25}               
		SSPL (w/ PCM)    									&VGG-Sound10k   	&\textbf{0.314}        &\textbf{0.369} \\
		\midrule
		Attention \cite{Senocak18}$_{\text{CVPR}18}^{*}$    &VGG-Sound144k    	&0.171                 &0.287 \\
		AVObject \cite{Afouras20}$_{\text{ECCV}20}$  	    &VGG-Sound144k    	&0.297 			       &0.357 \\
		HardWay \cite{Chen21c}$_{\text{CVPR}21}^{*}$	    &VGG-Sound144k    	&0.319	   			   &0.370 \\
		\rowcolor{lightgray!25}
		SSPL (w/o PCM)   									&VGG-Sound144k    	&0.270	   			   &0.348 \\
		\rowcolor{lightgray!25}
		SSPL (w/ PCM)    									&VGG-Sound144k    	&\textbf{0.339}	   	   &\textbf{0.380} \\
		\bottomrule
	\end{tabular}%
	\vspace{-2.5mm}
	\caption{\textbf{Quantitative localization results on VGG-SS test set.}}%
	\label{tab:compare_vggss}
	\vspace{-2mm}
\end{table}

\begin{figure}
	\centering
	\includegraphics[width=0.48\linewidth]{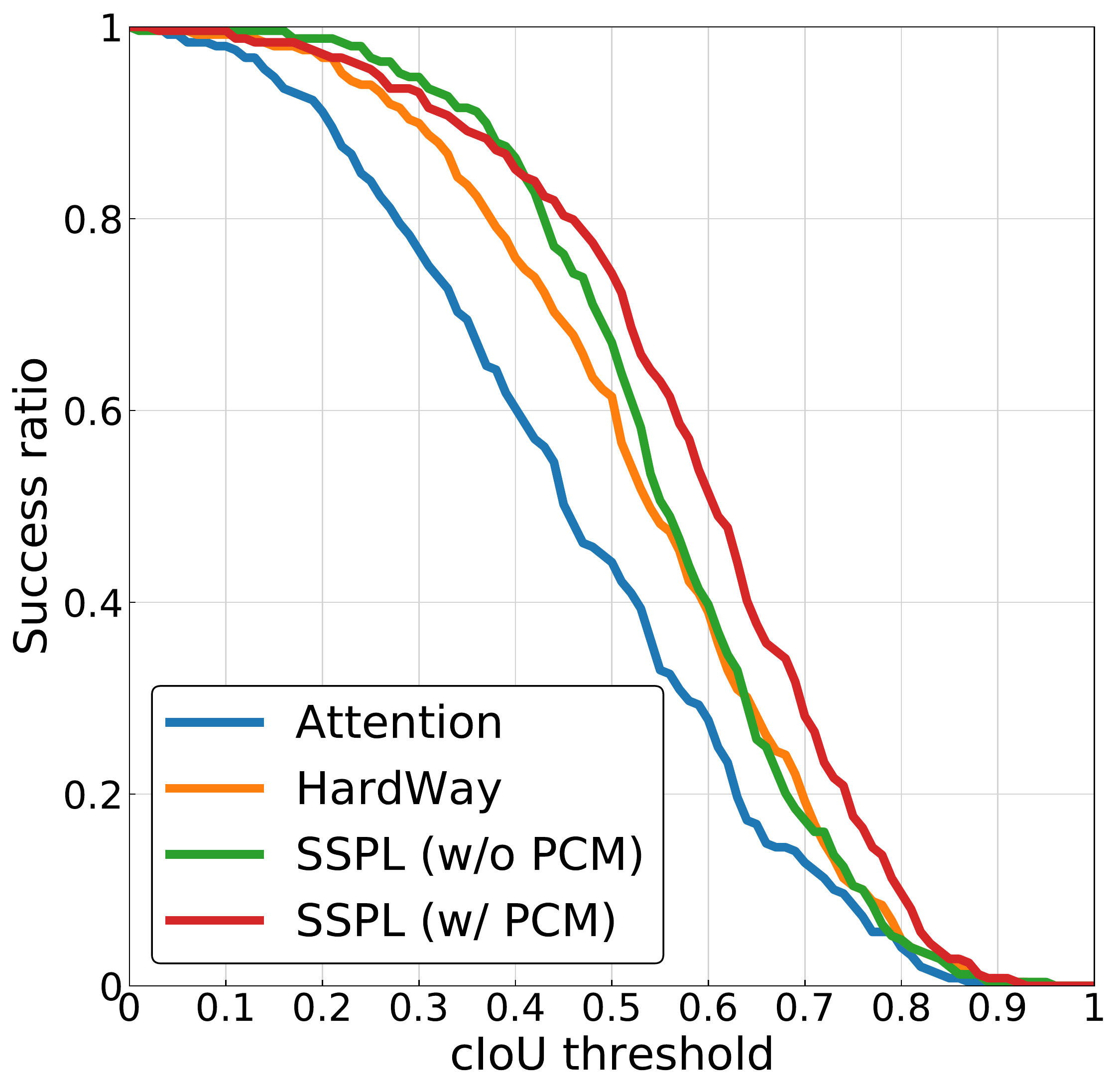}\hspace{2mm}
	\includegraphics[width=0.48\linewidth]{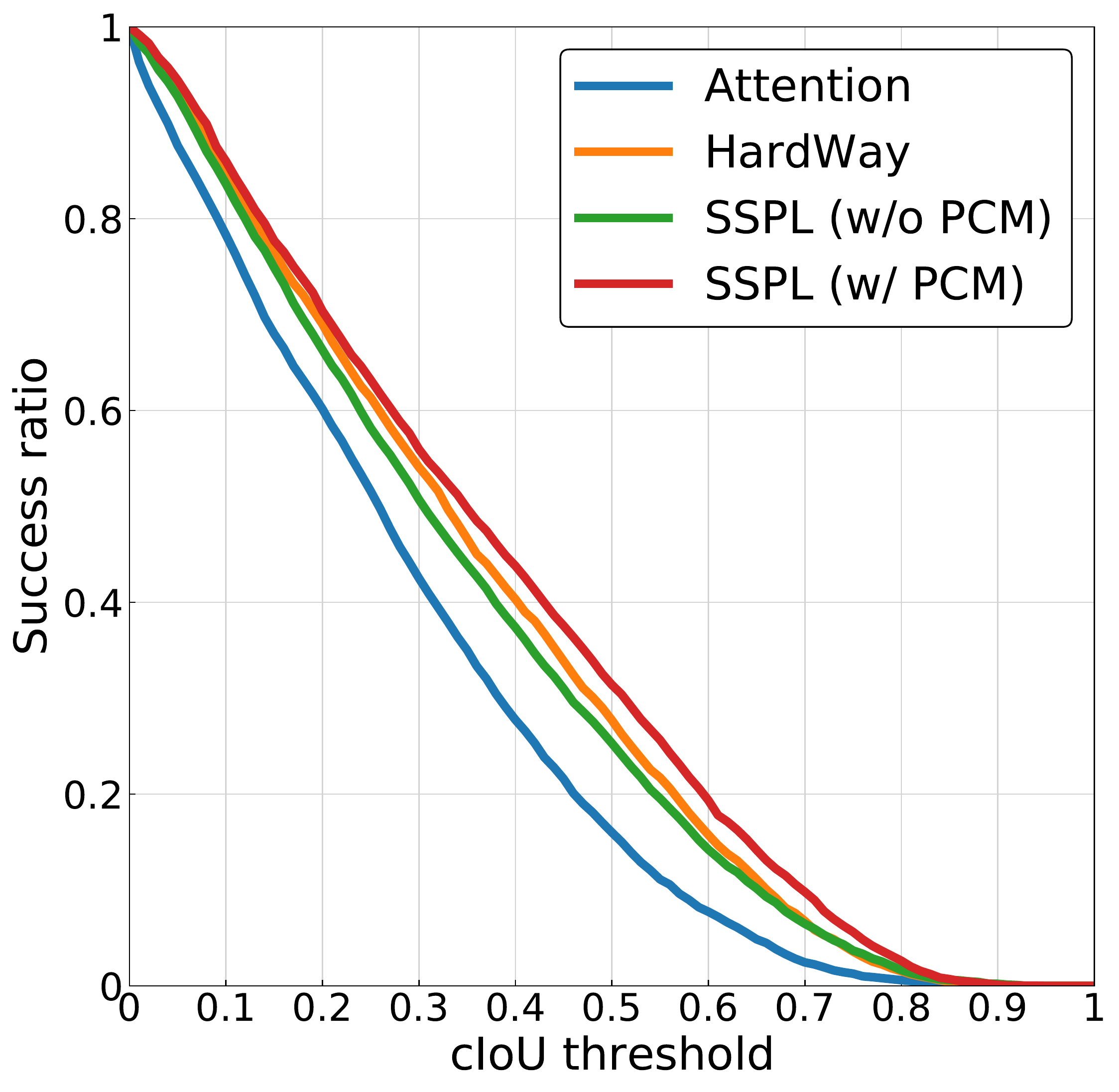}%
	\vspace{-3mm}
	\caption{\textbf{Success ration with varying cIoU thresholds.} \textbf{Left:} SoundNet-Flickr test set. \textbf{Right:} VGG-SS test set. Best viewed in color and by zooming in.}%
	\label{fig:success_rat_vs_ciou_thres}
	\vspace{-3mm}
\end{figure}

To address diverse demands for sound localization fineness, we compute cIoU scores with various thresholds as shown in \Cref{fig:success_rat_vs_ciou_thres}. The proposed method, SSPL (w/ PCM), again consistently surpasses the state-of-the-art (HardWay) under all thresholds.

\subsection{Qualitative Analysis}\label{sec:qualit_analysis}
\begin{figure*}
	\centering
	\begin{subfigure}{0.49\linewidth}
		\includegraphics[width=\linewidth]{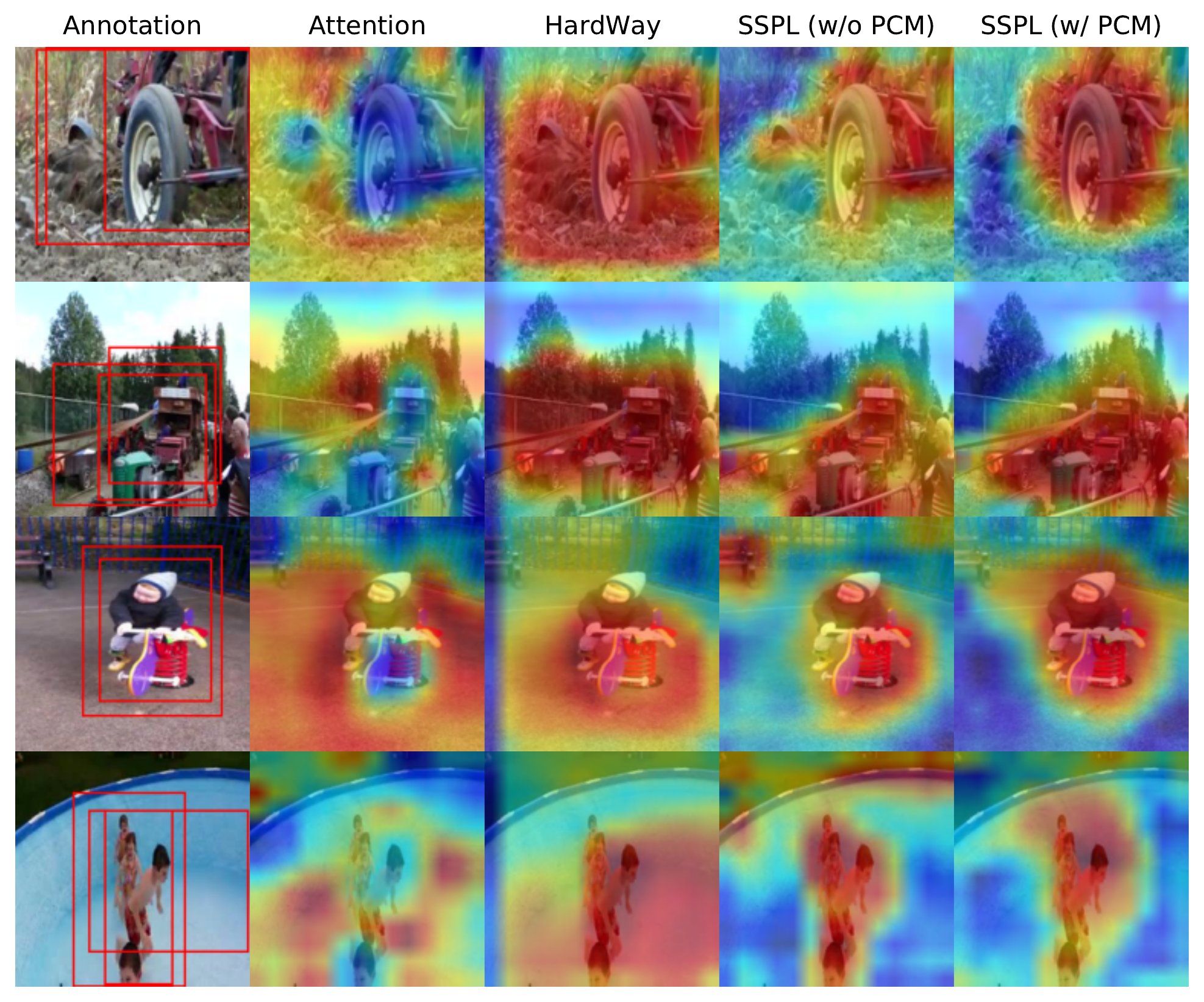}
		\vspace{-5mm}
		\caption{Visualisation on SoundNet-Flickr test set}
		\label{fig:vis_att_map_flickr}
	\end{subfigure}
	\hspace{0.2mm}
	\begin{subfigure}{0.49\linewidth}
		\includegraphics[width=\linewidth]{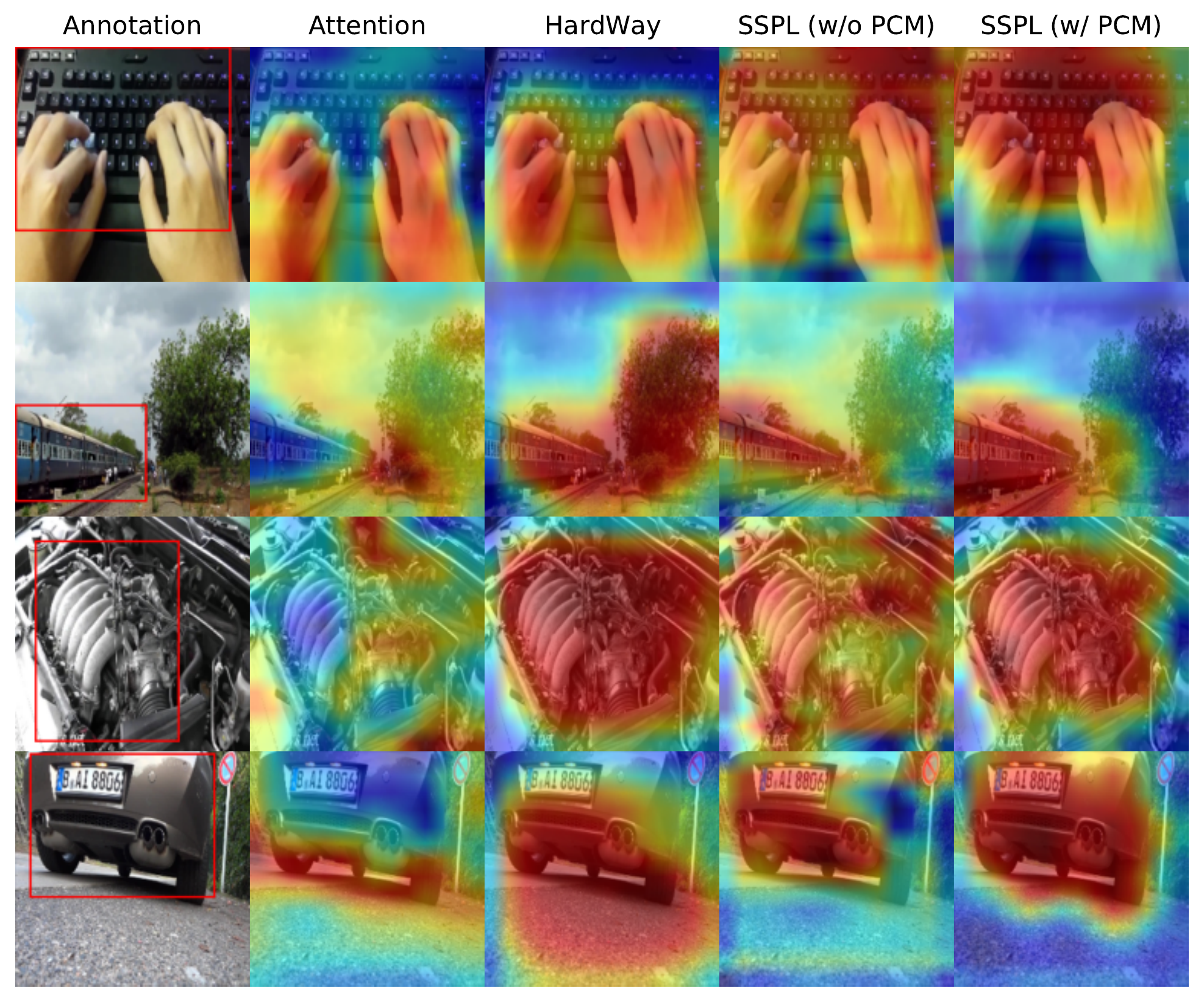}
		\vspace{-5mm}
		\caption{Visualisation on VGG-SS test set}
		\label{fig:vis_att_map_vggss}
	\end{subfigure}%
	\vspace{-3mm}
	\caption{\textbf{Qualitative comparisons.} In each panel, the first column shows images accompanied with annotations, and remaining columns represent the predicted localization of sounding objects. Here the attention map or similarity map produced by different methods is visualized as the localization map. Note that for SoundNet-Flickr the bounding boxes are derived from multiple annotators.}%
	\label{fig:vis_att_map}
	\vspace{-1mm}
\end{figure*}
We provide visualized localization results in \Cref{fig:vis_att_map}. We observe that Attention \cite{Senocak18} is prone to overlook target objects (\eg, the first and second rows in \Cref{fig:vis_att_map_flickr}) and cover unrelated background details (\eg, ground and sky). Since localization map also visualizes similarities between audio and visual features, the inaccurate localization indicates that Attention (random sampling of negatives) has the potential to misalign features. Although HardWay \cite{Chen21c} presents more centralized attention via hard negative mining, it easily underestimates (\eg, the first row in \Cref{fig:vis_att_map_vggss}) or overestimates (\eg, the second row in \Cref{fig:vis_att_map_vggss}) extents of sounding objects. This is probably because positive and negative regions in different images cannot be simply distinguished by the same thresholding parameters \cite{Chen21c}. By contrast, our SSPL can cover the main region of interest, and the use of PCM further helps reduce the influence of background noise, leading to more accurate localization.

\subsection{Ablation Study}\label{sec:ablation_study}
In this section, we delve deeper into SSPL by conducting extensive ablation studies. Unless otherwise specified, all experiments are performed on SoundNet-Flickr dataset.

\begin{table}
	\tabcolsep=9.65pt
	\footnotesize
	\centering
	\begin{tabular}{*{6}{c}}
		\toprule
		{}            &Pre-train           &Stop-grad       &$T$         &cIoU $\uparrow$   &AUC $\uparrow$ \\
		\midrule\midrule
		(a) 	      &{}				   &{}			    &{}	 	     &0.141            &0.147 \\
		(b) 	      &$\checkmark$		   &{}			    &{}	 	     &0.382            &0.432 \\
		(c)  	      &{} 	               &$\checkmark$	&{}	 	     &0.570            &0.511 \\
		(d)   	   	  &$\checkmark$ 	   &$\checkmark$	&{}	 	     &\textbf{0.671}   &\textbf{0.556} \\
		\midrule
		(e)   	   	  &$\checkmark$ 	   &$\checkmark$ 	&1           &0.655            &0.562 \\
		(f)   	   	  &$\checkmark$ 	   &$\checkmark$ 	&3 		     &0.719            &0.584 \\
		(g)   	   	  &$\checkmark$ 	   &$\checkmark$ 	&5 		     &\textbf{0.743}   &\textbf{0.587} \\
		\bottomrule
	\end{tabular}%
	\vspace{-2.5mm}
	\caption{\textbf{Ablation on training strategies.} ``Pre-train'' represents whether using the ImageNet-pretrained backbone to extract visual features, and $T$ denotes the recursive cycles for iterative computing in PCM during training.}%
	\label{tab:ablation}
	\vspace{-3mm}
\end{table}
\noindent\textbf{Training Strategy.} As discussed in prior art \cite{Chen21a}, a simple Siamese network without using negative samples can easily suffer from the problem of representation collapse. In this regard we evaluate key factors of SSPL that facilitate audio-visual learning. In \Cref{tab:ablation}a we train the model from scratch while removing the stop-gradient operation, which indeed causes collapse in our practice. The variant with only pre-training strategy (\Cref{tab:ablation}b) improves performance because of the better parameter initialization, but it dose not avoid collapsed solution yet. Adding stop-gradient alone during training (\Cref{tab:ablation}c) can obtain obvious gains, and the combination with pre-training (\Cref{tab:ablation}d) further boosts cIoU to $0.671$, which is the default configuration of vanilla SSPL. 

Based on above configuration, we perform additional ablation on the recursive cycles for representation updates in PCM. The performance sightly drops by 2\% as conducting feedback and feedforward representation updates (\cref{eq:feedback_pred,eq:feedback_repres,eq:feedforward_error,eq:feedforward_repres}) only once (\Cref{tab:ablation}e). This is because one computing step is not enough for PCM to reduce prediction errors between audio and visual features, and such non-negligible errors could degrade the subsequent learning. However, by increasing recursive cycles, SSPL can harvest significant performance improvements (nearly 10\% in \Cref{tab:ablation}f and over 13\% in \Cref{tab:ablation}g, respectively).

In summary, the results demonstrate that stop-gradient also works in our audio-visual setting to prevent collapse; and that both pre-training and PCM induce the model to learn effectively so as to promote localization accuracy.

\noindent\textbf{Augmentation.} We investigate the influence of various image augmentations on localization. As shown in \Cref{tab:img_aug}, with the random crop baseline, our method can already achieve reasonable performance, indicating that object scales really matter in SSPL. However, except for horizontal flip (over 30\% and 8\% improvements on two datasets, respectively), randomly combining other augmentations with crop cannot obtain consistent gains. This is because compared with other combinations, the spatial augmentations (random crop + horizontal flip) are more suitable for the pretrained and frozen VGG \cite{Simonyan15} to extract semantic visual features. Since our work is inspired by SimSiam \cite{Chen21a}, we also adopt its data augmentation strategies in SSPL, but find no benefits in this setting. Therefore, in all experiments we take the spatial augmentations by default.
\begin{table}
	\tabcolsep=9.3pt
	\footnotesize
	\centering
	\begin{tabular}{l*{4}{c}}
		\toprule
		\multirow{2}{*}{Augmentation}   &\multicolumn{2}{c}{SoundNet-Flickr}       &\multicolumn{2}{c}{VGG-SS} \\
		\cmidrule(lr){2-3}\cmidrule(lr){4-5}
		&cIoU $\uparrow$       &AUC $\uparrow$     &cIoU $\uparrow$        &AUC $\uparrow$ \\
		\midrule\midrule
		Crop (baseline)         		&0.514				   &0.499              &\underline{0.233}	   &0.324 \\
		\midrule
		+ Horizontal flip				&\underline{0.671}	   &\underline{0.556}  &\textbf{0.253}		   &\textbf{0.335} \\
		+ Vertical flip					&0.667				   &0.551              &0.213				   &0.317 \\
		+ Translation					&0.643				   &0.541              &0.216				   &0.313 \\
		+ Rotation						&0.639				   &0.543              &0.227				   &\underline{0.331} \\
		+ Grayscale						&0.610				   &0.535              &0.226				   &0.318 \\
		+ Color jittering               &\textbf{0.679}		   &\textbf{0.560}     &0.232				   &0.328 \\
		+ Gaussian blur					&0.619				   &0.533              &0.204				   &0.299 \\
		\bottomrule
	\end{tabular}%
	\vspace{-2.5mm}
	\caption{\textbf{Ablation on image augmentations.} \textbf{Bold} indicates the best and \underline{Underline} the runner-up. Parameters used to generate different augmentations are provided in supplement.}%
	\label{tab:img_aug}
	\vspace{-1.5mm}
\end{table}

\noindent\textbf{Scaling Method in AM.} The similarity map takes values in $[-1,1]$ and is adapted to weigh visual features in AM. Here we study different methods that can scale similarity range into $[0,1]$. ReLU is used in \cite{Qian20} to compact the similarities less than 0, but in our model enforcing those negative values to be equal produces worst results, as shown in \Cref{tab:scale_method}. While sigmoid and softmax \cite{Senocak18} boost performance by taking all different similarities into account, they shrink values into a proper subset of $[0,1]$. The min-max normalization (\cref{eq:norm_sim_map}), by contrast, takes a step forward and separates minima and maxima to the largest extent, yielding best results among others. This reveals that the relative importance between spatial-wise visual features is more crucial than the feature value per se for sound localization task.
\begin{table}
	\tabcolsep=12pt
	\footnotesize
	\centering
	\begin{tabular}{l*{2}{c}}
		\toprule
		Scaling method          		 &cIoU $\uparrow$      &AUC $\uparrow$ \\
		\midrule\midrule
		ReLU \cite{Qian20}      		 &0.353				   &0.424 \\
		Sigmoid				    		 &0.647				   &0.547 \\
		Softmax \cite{Senocak18}		 &0.667				   &0.554 \\
		ReLU + Softmax \cite{Senocak18}	 &0.574				   &0.531 \\
		Min-Max Norm.	        		 &\textbf{0.671}	   &\textbf{0.556} \\
		\bottomrule
	\end{tabular}%
	\vspace{-2.5mm}
	\caption{\textbf{Ablation on scaling methods.}}%
	\label{tab:scale_method}
	\vspace{-3mm}
\end{table}

\begin{figure}
	\centering
	\includegraphics[width=0.48\linewidth]{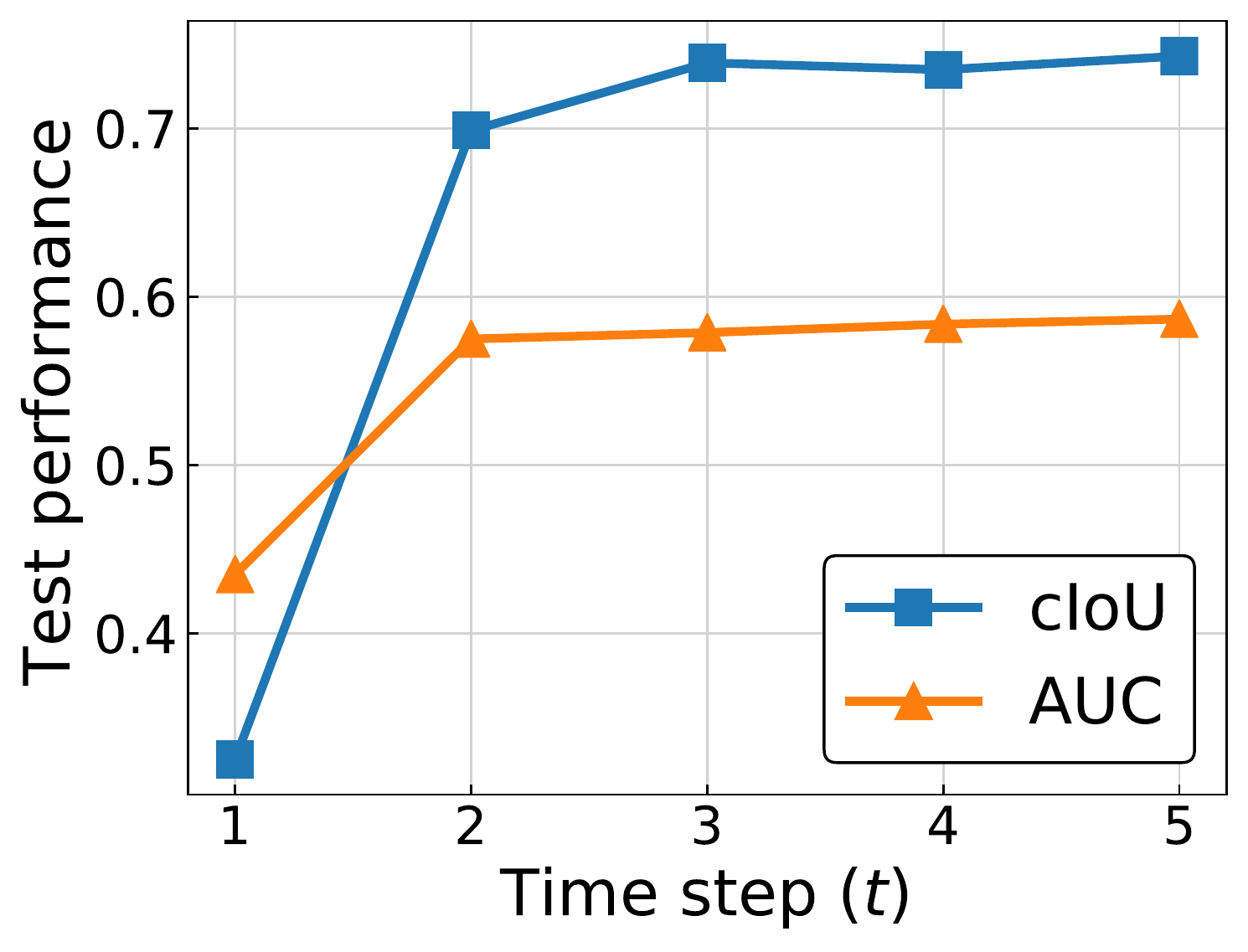}\hspace{2mm}
	\includegraphics[width=0.48\linewidth]{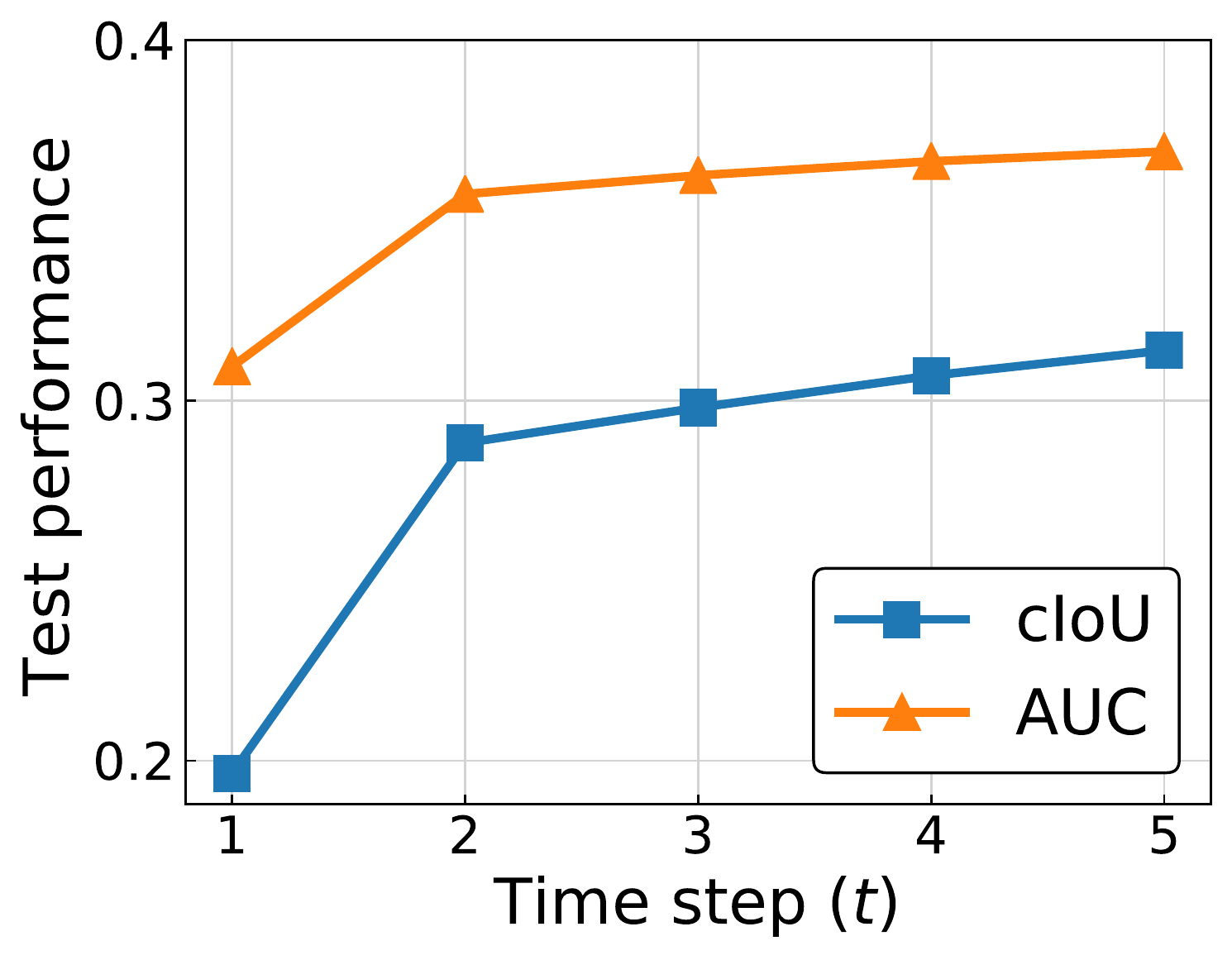}%
	\vspace{-3mm}
	\caption{\textbf{Performance with PCM's iterations during testing.} \textbf{Left:} SoundNet-Flickr test set. \textbf{Right:} VGG-SS test set.}%
	\label{fig:perform_vs_cycles}
	\vspace{-3mm}
\end{figure}

\begin{figure}
	\centering
	\includegraphics[width=\linewidth]{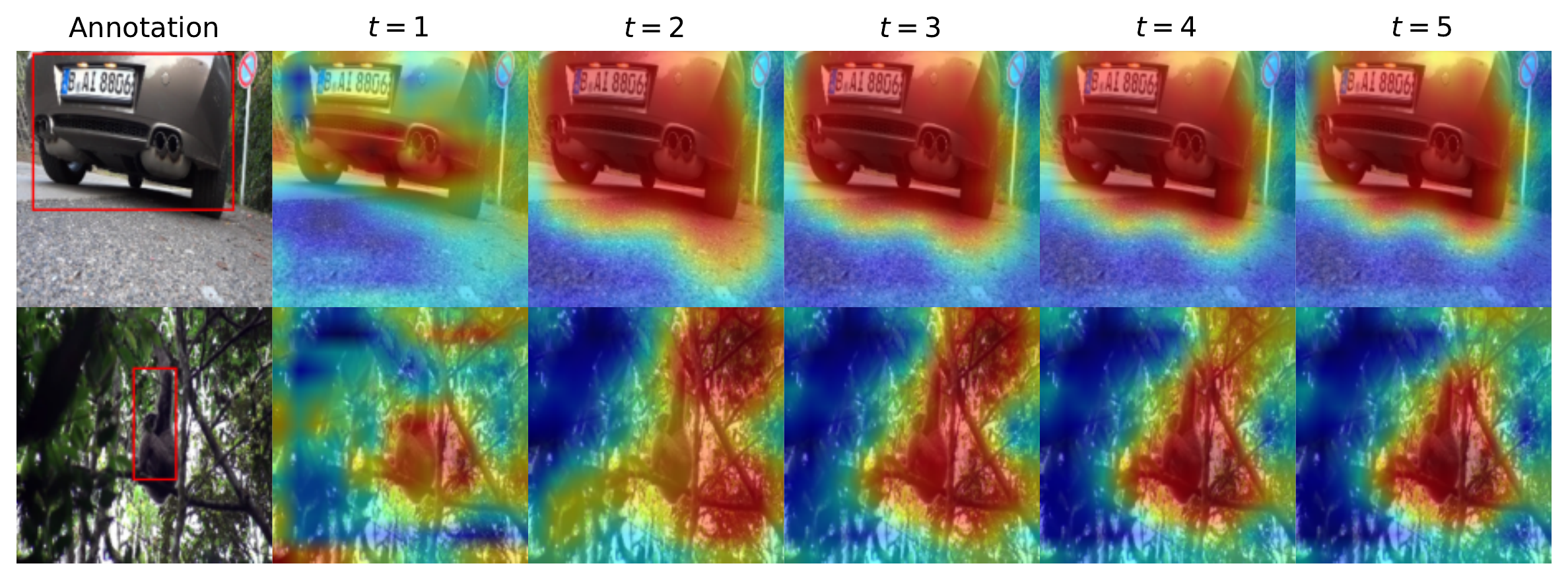}%
	\vspace{-3mm}
	\caption{\textbf{Attention map with PCM's iterations during testing.} Illustrations are from VGG-SS test set.}%
	\label{fig:vis_att_map_vs_iters}
	\vspace{-3mm}
\end{figure}

\noindent\textbf{Further Analysis of PCM.} We empirically clarify the remarkable ability of PCM to boost sound localization. Since PCM features an iterative computing procedure, we inspect the performance of SSPL (w/ PCM) with different iterations in \Cref{fig:perform_vs_cycles}. We observe that the localization accuracy tends to increase given more iterative computations, especially at the initial three time steps. To understand why this is the case, we look into attention maps from some test samples, as shown in \Cref{fig:vis_att_map_vs_iters}. PCM infers different visual representations with varying time steps (1 through 5), which are further used by AM to yield different attention maps. Attention is less definitive (light red on sounding objects) and/or inaccurate (crimson on backgrounds) at early time steps. At later time steps, however, the model corrects itself to pay more definitive and accurate attention to the objects of interest. Adjusting attention in such a coarse-to-fine manner is particularly helpful to address ambiguous cases, where the object's appearance may be similar to backgrounds (\eg, the second row in \Cref{fig:vis_att_map_vs_iters}).

\section{Conclusion and Future Works}
In this work, we have developed a self-supervised audio-visual learning method, SSPL, that improves visual sound localization performance by explicit positive mining. A three-stream network, as well as its training strategy, was designed to explore correspondence between sound and video frame from the same video clip. We further proposed PCM to align audio and visual features via cross-modal feature prediction, which boosts localization accuracy significantly. Our approach shows promising performance on sound localization task, especially achieving the new state-of-the-art on SoundNet-Flickr benchmark. 

While SSPL excels at single sound source localization, it is not applicable to localize multiple sound sources in unconstrained videos \cite{Qian20}, which is still a challenge for the community. A potential solution is to develop weakly- or semi-supervised methods. We leave it for future works.

\vspace{1mm}
{\small\noindent\textbf{Acknowledgements.} This work was supported in part by the Major Project for New Generation of AI (No. 2018AAA0100400), in part by the National Natural Science Foundation of China (Nos. 61836014, U21B2042, 62072457, 62006231, and 61976174), and in part by the Project funded by China Postdoctoral Science Foundation (No. 2021M703489).}

{\small
	\bibliographystyle{ieee_fullname}
	\bibliography{mybib}
}

\newpage

\setcounter{section}{0}
\setcounter{figure}{0}
\setcounter{table}{0}
\setcounter{equation}{0}
\setcounter{algorithm}{0}

\renewcommand{\thesection}{\Alph{section}}
\renewcommand{\thefigure}{S\arabic{figure}}
\renewcommand{\thetable}{S\arabic{table}}
\renewcommand{\theequation}{S\arabic{equation}}
\renewcommand{\thealgorithm}{S\arabic{algorithm}}
\renewcommand{\algorithmiccomment}[1]{\hfill{\color{gray}$\triangleright$#1}}

\noindent\textbf{\Large Supplementary Material}

\vspace{5mm}
This supplementary material contains four parts:
\begin{itemize}
	\item \Cref{sec:pcm} presents full derivations to formulate the representation update rules of predictive coding module (PCM).
	
	\item \Cref{sec:implement_details} provides more details on our implementation.
	
	\item \Cref{sec:orig_vs_repreo} compares the localization performance of Attention \cite{Senocak18} and HardWay \cite{Chen21c} in original papers with our reproductions.
	
	\item \Cref{sec:add_qualit_results} gives additional ablation and visualisation results.
\end{itemize}

\section{Full Formulation of PCM}\label{sec:pcm}
\begin{figure}
	\centering
	\includegraphics[width=\linewidth]{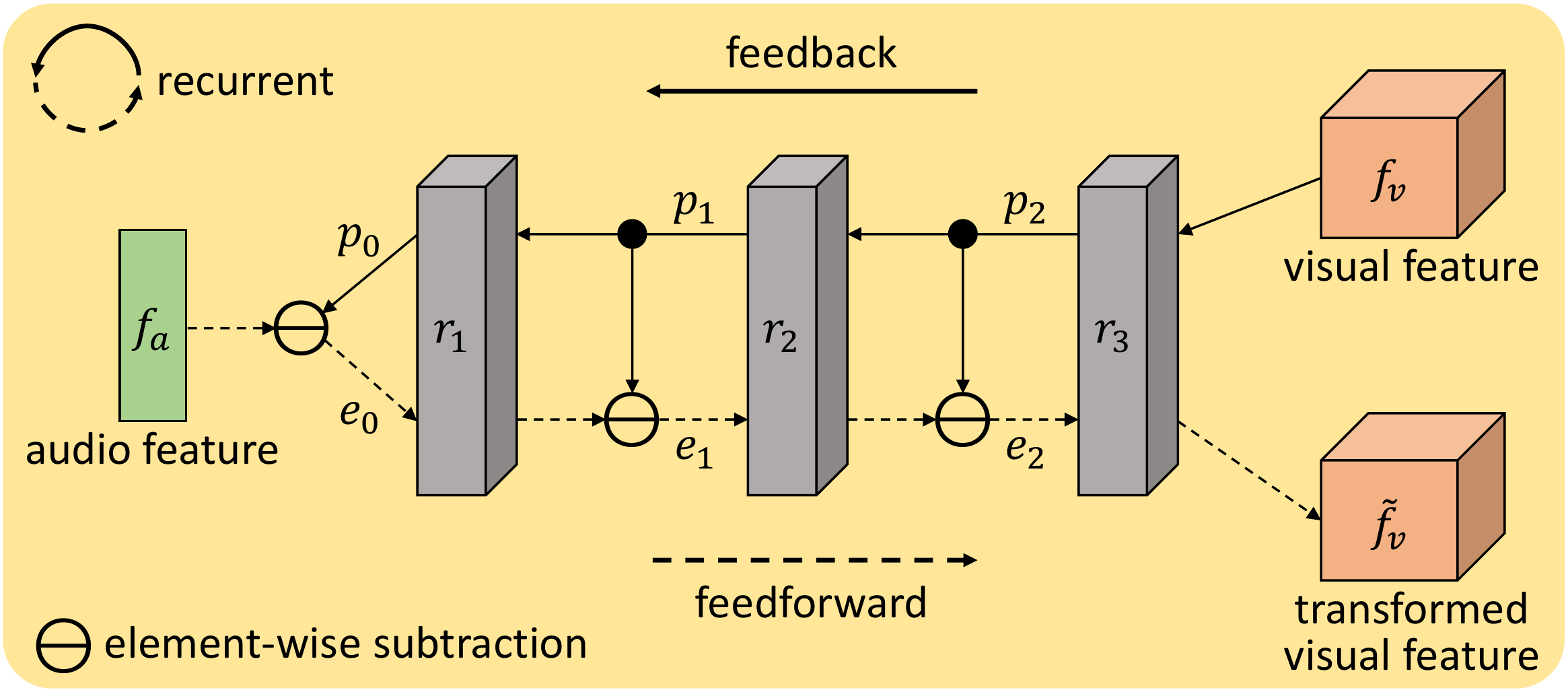}%
	\vspace{-1.5mm}
	\caption{\textbf{Overview of predictive coding module (PCM).} For simplicity we only show a 3-layer version.}%
	\label{fig:overview_pcm}
\end{figure}

The PCM, proposed for audio and visual feature alignment, plays an important role in improving sound localization performance of SSPL. As shown in \Cref{fig:overview_pcm}, the key idea underlying PCM consists of three parts: (1) a feedback process (solid line) updates representations with the top-down predictions that originate from the visual feature; (2) a feedforward process (dashed line) also updates representations but with the bottom-up prediction errors that evolve from the audio feature; (3) a recursive modulation mechanism works to conduct the two processes alternatively. In the following, we first formulate the optimization objective of PCM, and then derive the representation update rules of the two processes, respectively, which are followed by a brief summary and a formal algorithm. \emph{Note that for applications of PCM, we only need to explicitly update representations according to the rules given in \cref{eq:feedback_pred_s,eq:feedback_repres_s,eq:feedforward_error_s,eq:feedforward_repres_s}, without performing derivations again}.

Denote by $f_{a}$ the audio feature, by $f_{v}$ the visual feature, by $r_{l}(t), l\in\{1,\dots,L\}, t\in\{0,\dots,T\}$ the representation of the $l$-th layer of PCM network at time step $t$, and by $W_{l,l-1}$ the feedback connection weights from layer $l$ to layer $l-1$ (and vice versa for $W_{l-1,l}$). 

\vspace{2.5mm}
\noindent\textbf{Optimization Objective.} At layer $l$, PCM minimizes the following compound loss:
\begin{equation}\label{eq:loss_pcm}
	\mathcal{L}_{PCM}^{l} = \frac{\alpha_{l}}{2}\underbrace{||r_{l-1} - \mathcal{G}((W_{l,l-1})^{T}r_{l})||_{2}^{2}}_{\mathcal{L}_{1}^{l}} + \frac{\beta_{l}}{2}\underbrace{||r_{l} - p_{l}||_{2}^{2}}_{\mathcal{L}_{2}^{l}},
\end{equation}
where the function $\mathcal{G}$ corresponds to a generative process, $\alpha_{l}$ and $\beta_{l}$ are scalars that control the weights of the two loss terms $\mathcal{L}_{1}^{l}$ and $\mathcal{L}_{2}^{l}$, and $p_{l}=\mathcal{G}((W_{l+1,l})^{T}r_{l+1})$ is the prediction of $r_{l}$.

Given the lower-level representation $r_{l-1}$ and the top-down prediction $p_{l}$, our goal is to estimate $r_{l}$ so as to decrease the loss in \cref{eq:loss_pcm}. Minimizing $\mathcal{L}_{1}^{l}$ w.r.t. $r_{l}$ leads to the representation that can be used to predict the \emph{lower level} of representation $r_{l-1}$, while minimizing $\mathcal{L}_{2}^{l}$ w.r.t. $r_{l}$ yields the representation that approximates the prediction signal $p_{l}$ coming from a \emph{higher level}. Therefore, the representation $r_{l}$ associates lower- and higher-level information by reducing two prediction errors in $\mathcal{L}_{1}^{l}$ and $\mathcal{L}_{2}^{l}$. Minimizing losses at all layers can implicitly drive predictions at different levels to be mutually consistent \cite{Spratling17}.

\vspace{1mm}
\noindent\textbf{Feedback Process.} This process acts to update representations based on predictions from higher levels. Following \cite{Wen18}, we set $\mathcal{G}(x)=x$, and then employ gradient descent to minimize $\mathcal{L}_{2}^{l}$ w.r.t. $r_{l}$, resulting in update rules:
\begin{align}
	p_{l}(t) &= (W_{l+1,l})^{T}r_{l+1}(t), \\
	\frac{\partial \mathcal{L}_{2}^{l}}{\partial r_{l}(t)} &= 2(r_{l}(t) - p_{l}(t)), \\
	r_{l}(t+1) &= r_{l}(t) - \eta_{l}\frac{\beta_{l}}{2}\frac{\partial \mathcal{L}_{2}^{l}}{\partial r_{l}(t)} \nonumber\\
	&= (1 - \eta_{l}\beta_{l})r_{l}(t) + \eta_{l}\beta_{l}p_{l}(t), \label{eq:feedback_repres0_s}
\end{align}
where $\eta_{l}$ is a non-negative scalar governing learning. For simplicity, let $b_{l}=\eta_{l}\beta_{l}$, and then \cref{eq:feedback_repres0_s} is rewritten as follows:
\begin{equation}\label{eq:feedback_repres1_s}
	r_{l}(t+1) = (1 - b_{l})r_{l}(t) + b_{l}p_{l}(t).
\end{equation}
PCM carries out the feedback updating from top layer $L$ to bottom layer $1$, where the prediction of $r_{L}(t)$ at top layer is set as the visual feature, \ie, $p_{L}(t)\equiv f_{v}$.

\vspace{1mm}
\noindent\textbf{Feedforward Process.} This process works to update representations by using prediction errors from lower levels. For layer $l$, the lower-level prediction error $e_{l-1}$ is the difference between $r_{l-1}$ and $p_{l-1}$. We use gradient decent to minimize $\mathcal{L}_{1}^{l}$ w.r.t. $r_{l}$, leading to the following update rules:
\begin{align}
	e_{l-1}(t) &= r_{l-1}(t) - p_{l-1}(t), \\
	\frac{\partial \mathcal{L}_{1}^{l}}{\partial r_{l}(t)} &= -2W_{l,l-1}e_{l-1}(t), \\
	r_{l}(t+1) &= r_{l}(t) - \kappa_{l}\frac{\alpha_{l}}{2}\frac{\partial \mathcal{L}_{1}^{l}}{\partial r_{l}(t)} \nonumber\\
	&= r_{l}(t) + \kappa_{l}\alpha_{l}W_{l,l-1}e_{l-1}(t), \label{eq:feedforward_repres0_s}
\end{align}
where $\kappa_{l}$ is a non-negative scalar like $\eta_{l}$. We also set $a_{l}=\kappa_{l}\alpha_{l}$ for simplicity. Similar to \cite{Wen18}, we replace the feedback connection weights $W_{l,l-1}$ in \cref{eq:feedforward_repres0_s} with the transposed feedforward connection weights $(W_{l-1,l})^{T}$, and thus can endow PCM with more degrees of freedom to learn. Consequently the update rule in \cref{eq:feedforward_repres0_s} can be rewritten as a feedforward operation:
\begin{equation}\label{eq:feedforward_repres1_s}
	r_{l}(t+1) = r_{l}(t) + a_{l}(W_{l-1,l})^{T}e_{l-1}(t).
\end{equation}
In this process, PCM updates representations from bottom layer $1$ to top layer $L$, where we let $r_{0}(t)\equiv f_{a}$ and $p_{0}(t)=(W_{1,0})^{T}r_{1}(t)$.

\vspace{2mm}
\noindent\textbf{Summary and Algorithm.} So far we formulate PCM with the simple linear activation functions. To introduce non-linearity into PCM, a nonlinear activation function $\phi$ (\eg, ReLU \cite{Nair10} used in \cite{Wen18} or GELU \cite{Hendrycks16} used in this work) is applied to the above update \cref{eq:feedback_repres1_s,eq:feedforward_repres1_s}. By taking the recursive computing into account, we summarize the two processes as follows.

\noindent\emph{Nonlinear feedback process ($l=L,L-1,\dots,1$):}
\begin{align}
	p_{l}(t) &= (W_{l+1,l})^{T}r_{l+1}(t), \label{eq:feedback_pred_s}\\
	r_{l}(t) &\gets \phi((1 - b_{l})r_{l}(t-1) + b_{l}p_{l}(t)). \label{eq:feedback_repres_s}
\end{align}

\noindent\emph{Nonlinear feedforward process ($l=1,2,\dots,L$):}
\begin{align}
	e_{l-1}(t) &= r_{l-1}(t) - p_{l-1}(t), \label{eq:feedforward_error_s}\\
	r_{l}(t) &\gets \phi(r_{l}(t) + a_{l}(W_{l-1,l})^{T}e_{l-1}(t)). \label{eq:feedforward_repres_s}
\end{align}

The two processes are conducted alternatively such that all representations in PCM are refined progressively. Finally, we transform the top layer representation at last time step, $r_{L}(T)$, to a new visual feature, $\tilde{f}_{v}$, with dimension the same as $f_{v}$ by a $1\times1$ convolution. The representation learning of SSPL can proceed based on this $\tilde{f}_{v}$, instead of $f_{v}$ as used in the vanilla SSPL. We present main computing steps of PCM in \cref{alg:update_repres}.
\begin{algorithm}[t]
	\caption{Update Representations in PCM}\label{alg:update_repres}
	\begin{algorithmic}[1]
		\renewcommand{\algorithmicrequire}{\textbf{Input:}}
		\renewcommand{\algorithmicensure}{\textbf{Output:}}
		\Require $f_{v}$ and $f_{a}$
		\Ensure $\tilde{f}_{v}$
		\vspace{1mm}
		\For {$t=0$ to $T$}
		\If {$t=0$}
		\State initialize representations
		\EndIf
		
		\For {$l=L$ to $1$}\Comment{feedback process}
		\If {$l=L$}
		\State $p_{l}(t) = f_{v}$
		\Else
		\State compute prediction $p_{l}(t)$: \cref{eq:feedback_pred_s}
		\EndIf
		\State update representation $r_{l}(t)$: \cref{eq:feedback_repres_s}
		\EndFor
		
		\For {$l=1$ to $L$}\Comment{feedforward process}
		\If {$l=1$}
		\State $e_{l-1}(t) = f_{a} - \phi((W_{l,l-1})^{T}r_{l}(t))$
		\Else
		\State obtain prediction error $e_{l-1}(t)$: \cref{eq:feedforward_error_s}
		\EndIf
		\State update representation $r_{l}(t)$: \cref{eq:feedforward_repres_s}
		\EndFor
		\EndFor
		
		\State $\tilde{f}_{v}=\texttt{Conv}_{1\times1}(r_{L}(T))$\Comment{transformed feature}
	\end{algorithmic}
\end{algorithm}

\section{Implementation Details}\label{sec:implement_details}
\subsection{Architecture of PCM}
For the feedback process of PCM, we use convolution layers ($\texttt{kernel\_size}=3$, $\texttt{stride}=1$, $\texttt{padding}=1$) followed by max pooling operation to reduce the spatial dimensionality of feature maps, while using $1\times1$ convolutions to decrease the number of channels. As for the feedforward process, the transposed convolutions (a.k.a. deconvolutions) are utilized and feature maps are upsampled by the ``bilinear'' upsampling algorithm, provided in PyTorch. Besides, the number of convolution layers is $L=3$. From top layer $L$ to bottom layer 1, the number of filters within each layer is 512, 512, and 128, respectively. The transposed convolution layers have the same setting. Moreover, we use GELU \cite{Hendrycks16} as the nonlinear activation function for both processes. To stabilize and accelerate training, we adopt the batch normalization \cite{Laurent16} before every non-linearity at each layer and at each time step, except the prediction of audio feature at bottom layer.

\subsection{Training Details for SSPL}\label{sec:train_setting}
The AdamW \cite{Loshchilov19} optimizer is employed to train our model, where we set $(\beta_{1}, \beta_{2})=(0.9, 0.999)$ and set weight decay to $10^{-4}$. In practice, we find that better performance could be achieved if the learning rate for projection and prediction MLPs is greater than that for remaining model parts. We show detailed learning rate settings in \Cref{tab:lrs}. During training, there are 256 image-audio pairs in each minibatch, which are distributed in parallel on 2 or 4 NVIDIA GeForce GTX 1080 Ti GPUs.
\begin{table}
	\tabcolsep=6.65pt
	\footnotesize
	\centering
	\begin{tabular}{lcccc}
		\toprule
		\multirow{2}{*}{Training set}   &\multicolumn{2}{c}{SSPL (w/o PCM)} &\multicolumn{2}{c}{SSPL (w/ PCM)} \\
		\cmidrule(lr){2-3}\cmidrule(lr){4-5}
		&$lr_{1}$        &$lr_{2}$          &$lr_{1}$        &$lr_{2}$ \\
		\midrule\midrule
		SoundNet-Flickr                 &$2\cdot10^{-3}$ &$5\cdot10^{-4}$   &$5\cdot10^{-5}$ &$2\cdot10^{-5}$ \\
		VGG-Sound						&$1\cdot10^{-2}$ &$5\cdot10^{-3}$   &$5\cdot10^{-5}$ &$2\cdot10^{-5}$ \\
		\bottomrule
	\end{tabular}%
	\vspace{-2.5mm}
	\caption{\textbf{Learning rate settings.} $lr_{1}$ denotes the learning rate for projection and prediction MLPs, $lr_{2}$ for remaining model parts.}%
	\label{tab:lrs}
\end{table}

\subsection{Image Augmentations for SSPL}\label{sec:aug_params}
As shown in \Cref{tab:aug_params}, a total of 8 image augmentations are considered in our method. We follow HardWay \cite{Chen21c} to select and set the first two augmentations: cropping with $224\times224$ resizing and horizontal flip. Then, we verify the effectiveness of other three spatial augmentations that are widely used in self-supervised visual representation learning \cite{Gidaris18,Ki21}, \ie, vertical flip, translation, and rotation. Additionally, since our work draws inspiration from SimSiam \cite{Chen21a}, we also take into account its augmentation strategies: grayscale, color jittering, and Gaussian blur, while keeping their settings the same as SimSiam.
\begin{table}
	\tabcolsep=9.8pt
	\footnotesize
	\centering
	\begin{tabular}{rl}
		\toprule
		Augmentation                 		&Parameter \\
		\midrule\midrule
		\multirow{4}{*}{Crop}		 		&$p=1$ \\
		&output size of \texttt{Resize} $=\texttt{int}(224 \times 1.1)$ \\
		&interpolation method of \texttt{Resize} $=$ BICUBIC \\
		&crop size $=224$ \\
		\midrule
		Horizontal flip				 		&$p=0.5$  \\
		\midrule
		Vertical flip				 		&$p=0.5$  \\
		\midrule
		\multirow{2}{*}{Translation} 		&$p=1.0$  \\
		&maximum absolute fraction $=(0.2, 0.2)$ \\
		\midrule
		\multirow{2}{*}{Rotation}	 		&$p=1.0$  \\
		&angle $\in \{0, 90, 180, 270\}$ \\
		\midrule
		Grayscale	 				 		&$p=0.2$  \\
		\midrule
		\multirow{5}{*}{Color jittering}    &$p=0.8$  \\
		&maximum brightness adjustment $=0.4$ \\
		&maximum contrast adjustment $=0.4$ \\
		&maximum saturation adjustment $=0.4$ \\
		&maximum hue adjustment $=0.1$ \\
		\midrule									
		\multirow{2}{*}{Gaussian blur} 	    &$p=0.5$  \\
		&$\sigma \in [0.1, 2.0)$ \\
		\bottomrule
	\end{tabular}%
	\vspace{-2.5mm}
	\caption{\textbf{Parameters used to generate image augmentations.} $p$ denotes the probability that the corresponding operation will be performed.}%
	\label{tab:aug_params}
\end{table}

\section{Reproducing Related Methods}\label{sec:orig_vs_repreo}
The quantitative comparisons in the main text are partially based on our reproductions of two related methods: Attention \cite{Senocak18} and HardWay \cite{Chen21c}. We reimplement them as faithfully as possible by following each corresponding paper. As show in \Cref{tab:orig_vs_repro_flickr}, we are able to improve these two methods on SoundNet-Flickr by small and straightforward modifications. Specifically, we use Crop and Horizontal flip given in \Cref{tab:aug_params} to spatially augment images for Attention (\vs originally $320\times320$ resizing), same as HardWay and our method. We also fine tune the learning rate and weight decay for these two competitors in order to achieve their best performance.

\Cref{tab:orig_vs_repro_vggss} compares our reproductions with original papers' results on VGG-SS. In this case our reproductions are slightly lower than the original counterparts. Note that we had tried to adjust various hyper-parameters for HardWay training (\eg, learning rate, weight decay, batch size, and number of training epochs) for multiple times, but better performance than reproductions shown in \Cref{tab:orig_vs_repro_vggss} were not achieved. We contribute the performance discrepancy to the updated data in this benchmark. On the one hand, Chen \etal \cite{Chen21c} originally provides 5158 YouTube video IDs for testing, and users need to download, extract, and pre-process the designated audio and visual sources themselves. However, 466 videos (9\%) are not available (removed or prohibited download) at the time of conducting our experiments, leading to 4692 image-audio pairs for final testing. On the other hand, as clarified by the authors on the official project page\footnote{\href{https://github.com/hche11/Localizing-Visual-Sounds-the-Hard-Way}{https://github.com/hche11/Localizing-Visual-Sounds-the-Hard-Way}.}, some bounding box annotations are updated recently and consequently it could cause a 2\%-3\% difference on performance. Based on these two aspects, we think that our reproductions are reasonable.
\begin{table}
	\tabcolsep=7pt
	\footnotesize
	\centering
	\begin{tabular}{lclcc}
		\toprule
		Method    								    &Source  		&Training set  	 &cIoU $\uparrow$   &AUC $\uparrow$ \\
		\midrule\midrule
		\multirow{2}{*}{Attention \cite{Senocak18}} &O 	    &Flickr10k   	 &0.436             &0.449 \\
		&\cellcolor{lightgray!25}R 	    &\cellcolor{lightgray!25}Flickr10k     	 &\cellcolor{lightgray!25}0.442             &\cellcolor{lightgray!25}0.461 \\
		\multirow{2}{*}{HardWay \cite{Chen21c}} 	&O   	&Flickr10k     	 &0.582 			&0.525 \\
		&\cellcolor{lightgray!25}R 	    &\cellcolor{lightgray!25}Flickr10k     	 &\cellcolor{lightgray!25}0.615 			&\cellcolor{lightgray!25}0.535 \\
		\midrule
		\multirow{2}{*}{HardWay \cite{Chen21c}}     &O 		&Flickr144k  	 &0.699 			&0.573 \\
		&\cellcolor{lightgray!25}R 	    &\cellcolor{lightgray!25}Flickr144k	   	 &\cellcolor{lightgray!25}0.699				&\cellcolor{lightgray!25}0.590 \\
		\midrule
		\multirow{2}{*}{Attention \cite{Senocak18}} &O	 	&VGG-Sound10k	 &-                 &- \\
		&\cellcolor{lightgray!25}R 	    &\cellcolor{lightgray!25}VGG-Sound10k    &\cellcolor{lightgray!25}0.522             &\cellcolor{lightgray!25}0.502 \\
		\multirow{2}{*}{HardWay \cite{Chen21c}}     &O   	&VGG-Sound10k  	 &0.618 			&0.536 \\
		&\cellcolor{lightgray!25}R 	    &\cellcolor{lightgray!25}VGG-Sound10k  	 &\cellcolor{lightgray!25}0.647             &\cellcolor{lightgray!25}0.560 \\
		\midrule
		\multirow{2}{*}{HardWay \cite{Chen21c}}     &O   	&VGG-Sound144k   &0.719 			&0.582 \\
		&\cellcolor{lightgray!25}R 	    &\cellcolor{lightgray!25}VGG-Sound144k   &\cellcolor{lightgray!25}0.723	   			&\cellcolor{lightgray!25}0.605 \\
		\bottomrule
	\end{tabular}%
	\vspace{-2.5mm}
	\caption{\textbf{Our reproductions \vs original papers' results on SoundNet-Flickr test set.} ``O'' denotes results from original papers and ``R'' our reproductions.}%
	\label{tab:orig_vs_repro_flickr}
\end{table}

\begin{table}
	\tabcolsep=7pt
	\footnotesize
	\centering
	\begin{tabular}{lclcc}
		\toprule
		Method    					         		&Source  	&Training set       &cIoU $\uparrow$    &AUC $\uparrow$ \\
		\midrule\midrule
		\multirow{2}{*}{Attention \cite{Senocak18}} &$^{*}$     &VGG-Sound144k    	&0.185    			&0.302 \\
		&\cellcolor{lightgray!25}R     		&\cellcolor{lightgray!25}VGG-Sound144k    	&\cellcolor{lightgray!25}0.171              &\cellcolor{lightgray!25}0.287 \\
		\multirow{2}{*}{HardWay \cite{Chen21c}}		&O     		&VGG-Sound144k    	&0.344 				&0.382 \\
		&\cellcolor{lightgray!25}R 			&\cellcolor{lightgray!25}VGG-Sound144k    	&\cellcolor{lightgray!25}0.319	   			&\cellcolor{lightgray!25}0.370 \\
		\bottomrule
	\end{tabular}%
	\vspace{-2.5mm}
	\caption{\textbf{Our reproductions \vs original papers' results on VGG-SS test set.} ``$*$'' denotes results obtained from \cite{Chen21c}.}%
	\label{tab:orig_vs_repro_vggss}
\end{table}

\section{Additional Results}\label{sec:add_qualit_results}
\subsection{Ablation on Feature Fusion Methods}
In SSPL, visual and audio features are fused by the attention mechanism to compute audio-visual representation. Here we compare other three feature fusion methods, \ie, concatenation (Cat), multiplication ($\otimes$), and addition ($\oplus$), with our attention module (AM). We can see from \Cref{tab:feat_fuse_method} that our AM outperforms others by a large margin. This verifies efficacy of the attention-based feature interaction.
\begin{table}
	\tabcolsep=10pt
	\footnotesize
	\centering
	\begin{tabular}{*{5}{c}}
		\toprule
		Fusion method       &Cat			&$\otimes$      &$\oplus$     &AM (ours)\\
		\midrule\midrule
		cIoU $\uparrow$     &0.285            &0.538          &0.647 		&\textbf{0.671}\\
		AUC $\uparrow$      &0.414            &0.512          &0.540 		&\textbf{0.556}\\
		\bottomrule
	\end{tabular}%
	\vspace{-2.5mm}
	\caption{\textbf{Ablation on feature fusion methods.} We use different fusion methods in SSPL (w/o PCM), and train models on SoundNet-Flickr10k while evaluating on the standard benchmark.}%
	\label{tab:feat_fuse_method}
\end{table}

\subsection{Balance between Performance and Complexity of PCM}
In \Cref{tab:recursive_cycles}, we quantitatively compare performance and time complexity of SSPL with varying recursive cycles $T$. We find that more recursive cycles cannot always bring gains as performance tends to be saturated when $T>5$. Additionally, compared with SSPL (w/o PCM) that occupies 35.9 GFLOPs, SSPL (w/ PCM) conducts more operations with increasing $T$. As shown in \Cref{tab:recursive_cycles}, PCM takes, on average, 2.3 GFLOPs to complete one iteration. To balance between performance and time complexity, we set $T=5$ during training.
\begin{table}
	\tabcolsep=6.6pt
	\footnotesize
	\centering
	\begin{tabular}{*{7}{c}}
		\toprule
		$T$					&1		&3		&5		&6		&7		&8\\
		\midrule\midrule
		cIoU $\uparrow$		&0.655  &0.719  &0.743  &0.743  &0.759  &0.747\\
		AUC $\uparrow$		&0.562  &0.584	&0.587  &0.595  &0.595  &0.590\\
		GFLOPs $\downarrow$ &38.3   &43.0   &47.6   &49.9   &52.2   &54.5\\
		\bottomrule
	\end{tabular}%
	\vspace{-2.5mm}
	\caption{\textbf{Influence of recursive cycles $T$ in PCM.} All models are trained on SoundNet-Flickr10k and evaluated on the standard benchmark.}%
	\label{tab:recursive_cycles}
\end{table}

\subsection{Effect of False Negatives on Localization}
\begin{figure*}
	\centering
	\includegraphics[width=0.6\linewidth]{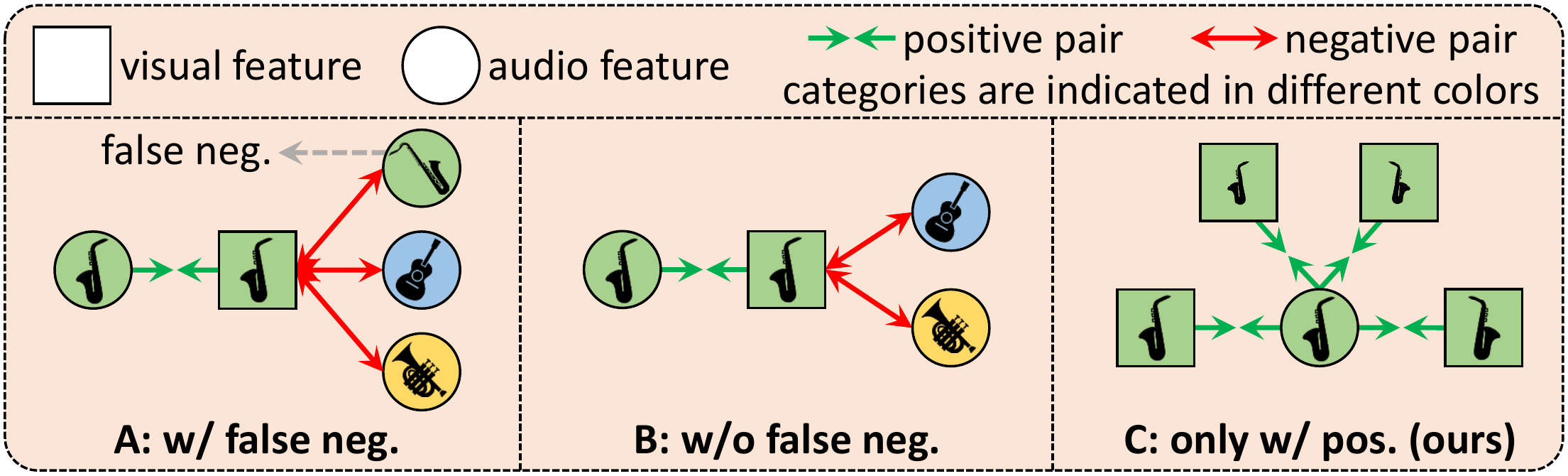}\\
	\vspace{1mm}
	\includegraphics[width=0.8\linewidth]{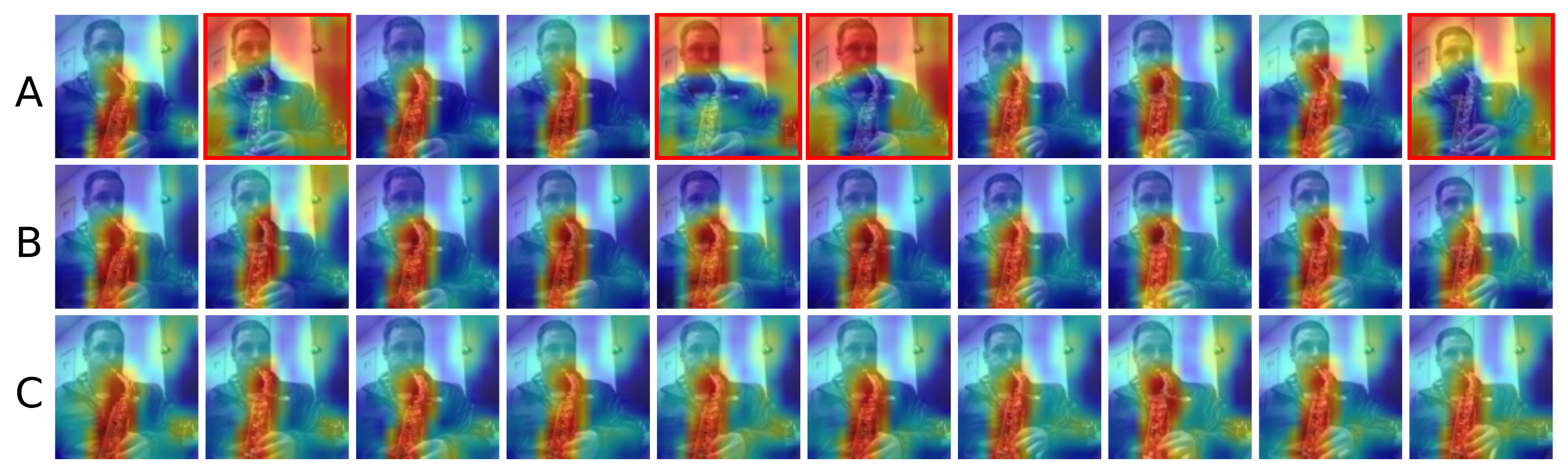}\\
	\vspace{1mm}
	\includegraphics[width=0.8\linewidth]{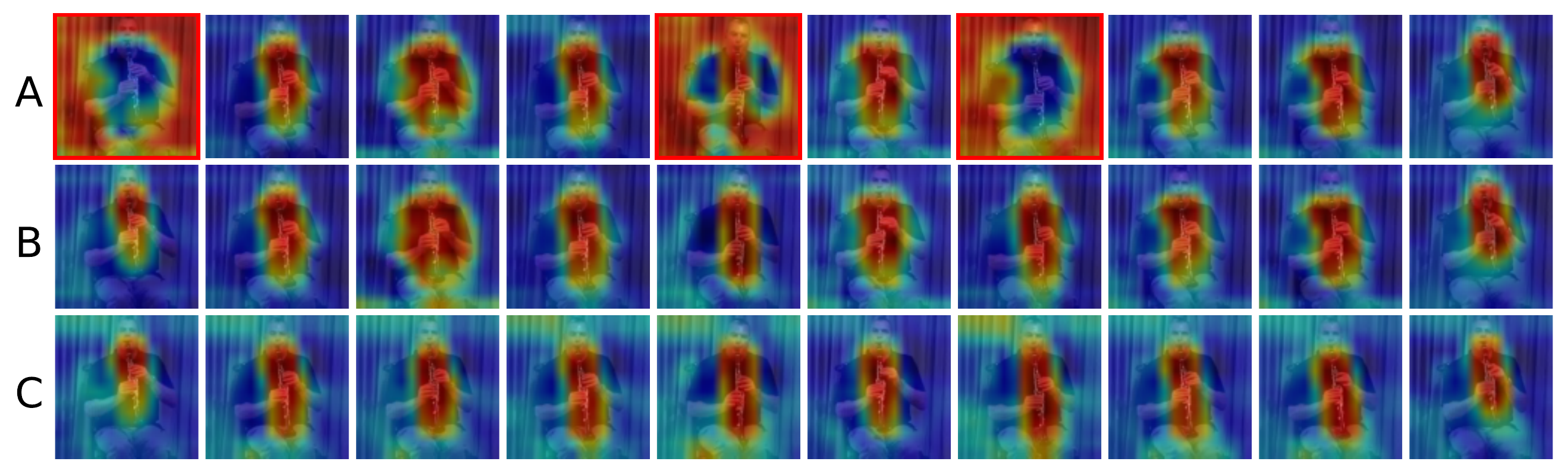}\\
	\vspace{1mm}
	\includegraphics[width=0.8\linewidth]{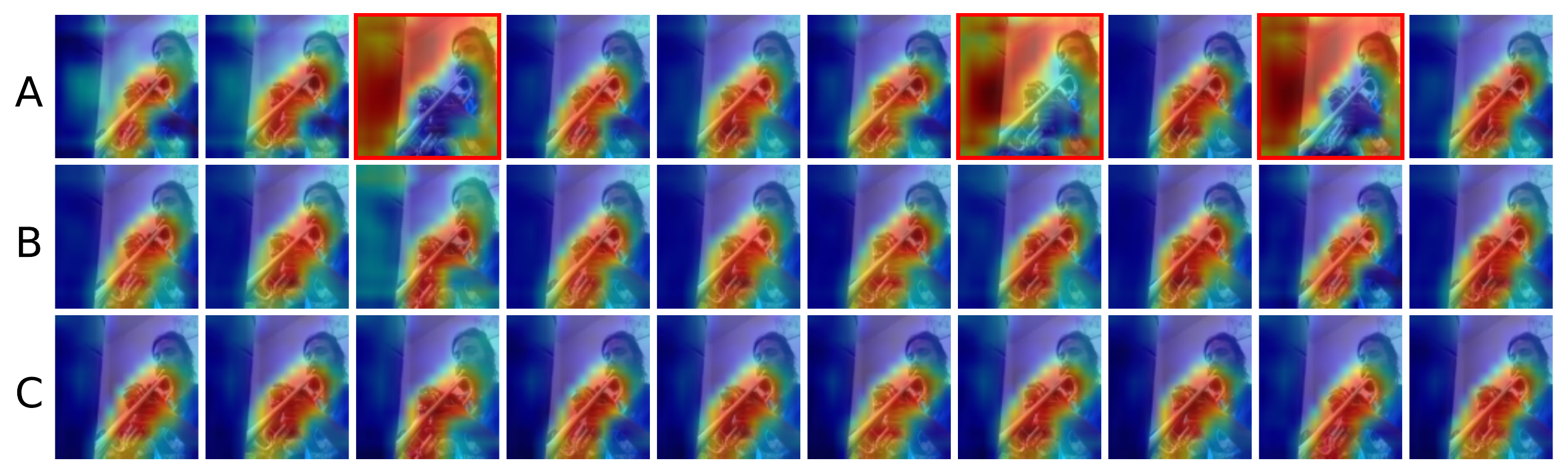}\\
	\vspace{1mm}
	\includegraphics[width=0.8\linewidth]{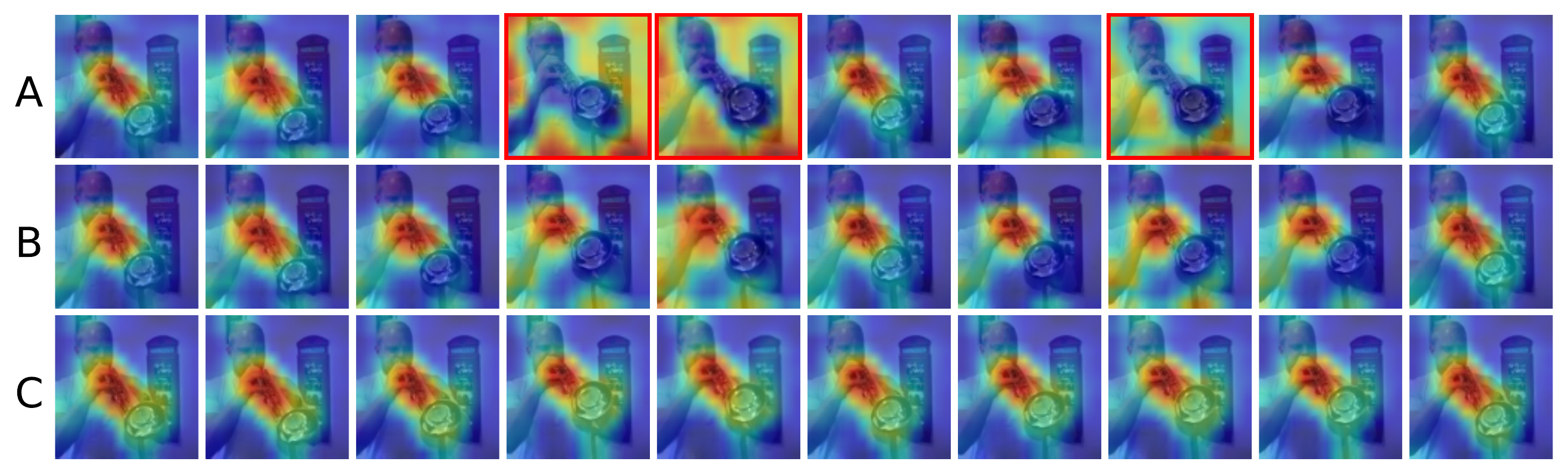}%
	\vspace{-2mm}
	\caption{\textbf{Visualisation of the effect of false negatives on sound localization.} \textbf{A} denotes the training strategy that uses both true and false negatives to perform constrastive learning; \textbf{B} indicates the method where false negatives do not take part in contrastive learning, but \emph{requiring class label to direct negative sampling}; \textbf{C} corresponds to our self-supervised method that only explores audio-visual positive pairs during learning. Here the false negatives are other videos' sounds that belong to the same category as the positive one. The images marked with red rectangle illustrate ambiguous localization results of method \textbf{A}. Models are trained on MUSIC \cite{Zhao18}.}%
	\label{fig:vis_effect_false_negs}
\end{figure*}

As discussed in the main text, learning with false negatives can induce ambiguity in localization results. In this section, we give more examples to empirically illustrate this effect. As shown in \Cref{fig:vis_effect_false_negs}, when the false negatives are allowed to take part in contrastive learning, sounding objects are easily ignored in final localization maps (method A). Although learning with true positive and negative samples harvests accurate localization, it requires class label to direct negative sampling (method B). By contrast, our method is able to obtain consistent localization among different image-audio pairs, without using negatives and labels at all (method C).

\subsection{Additional Qualitative Comparisons}
\begin{figure*}
	\centering
	\begin{subfigure}{0.49\linewidth}
		\includegraphics[width=\linewidth]{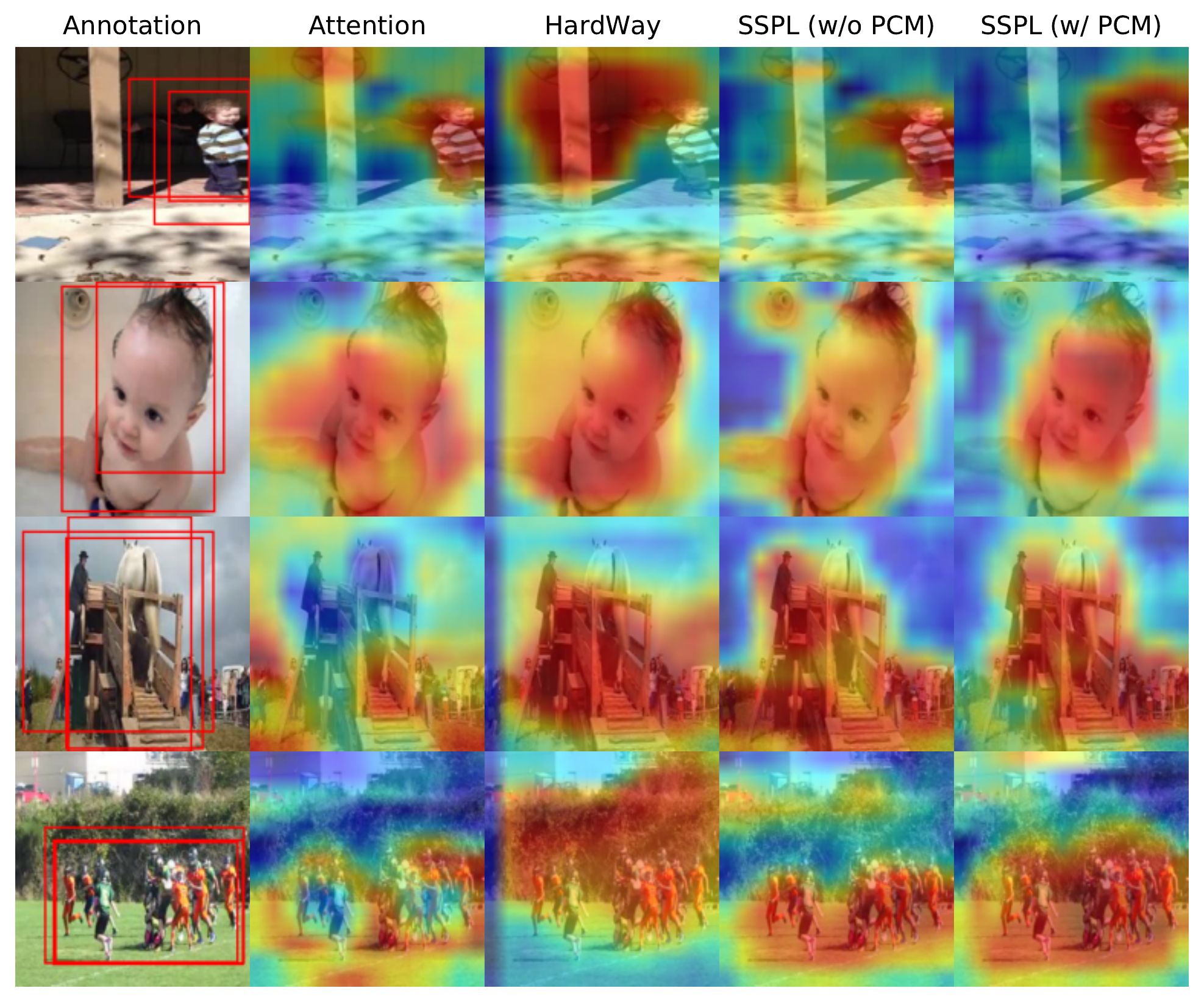}
		\caption{Visualisation on SoundNet-Flickr test set}
		\label{fig:vis_att_map_flickr_supp}
	\end{subfigure}
	\hspace{0.2mm}
	\begin{subfigure}{0.49\linewidth}
		\includegraphics[width=\linewidth]{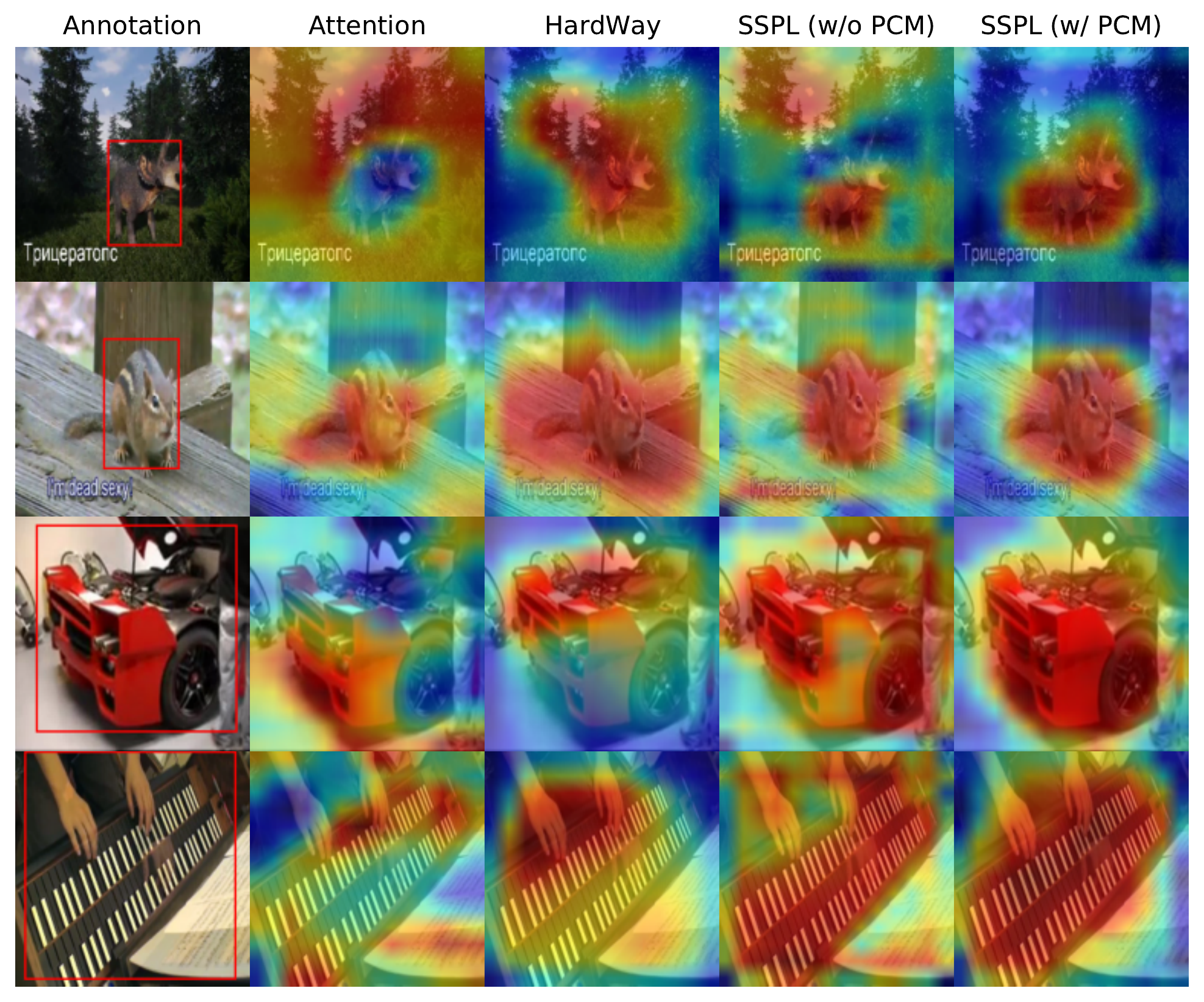}
		\caption{Visualisation on VGG-SS test set}
		\label{fig:vis_att_map_vggss_supp}
	\end{subfigure}%
	\vspace{-2mm}
	\caption{\textbf{Qualitative comparisons.} In each panel, the first column shows images accompanied with annotations, and remaining columns represent the predicted localization of sounding objects. Here the attention map or similarity map produced by different methods is visualized as the localization map. Note that for SoundNet-Flickr the bounding boxes are derived from multiple annotators.}%
	\label{fig:vis_att_map_supp}
\end{figure*}

In \Cref{fig:vis_att_map_supp}, we illustrate more localization examples from Attention \cite{Senocak18}, HardWay \cite{Chen21c}, and our method SSPL on two standard benchmarks: SoundNet-Flickr and VGG-SS. Qualitative evaluation results show that our method can localize the full extent of sounding objects, especially for SSPL (w/ PCM) that yields more accurate localization by ignoring background noise.

\end{document}